\documentclass[sn-mathphys,iicol,Numbered]{sn-jnl}

\usepackage{flushend}
\usepackage{graphicx}%
\usepackage{multirow}%
\usepackage{amsmath,amssymb,amsfonts}%
\usepackage{amsthm}%
\usepackage{mathrsfs}%
\usepackage[title]{appendix}%
\usepackage{colortbl}
\usepackage{xcolor}%
\usepackage{textcomp}%
\usepackage{manyfoot}%
\usepackage{booktabs}%
\usepackage{algorithm}%
\usepackage{algorithmicx}%
\usepackage{algpseudocode}%
\usepackage{listings}%
\usepackage{multirow}
\usepackage{makecell}
\usepackage{pifont}

\newlength\savewidth

\newcommand{\cmark}{\ding{51}}%
\newcommand{\xmark}{\ding{55}}%

\def\eg{\textit{e.g.}}
\def\ie{\textit{i.e.}}

\theoremstyle{thmstyleone}%
\theoremstyle{thmstyletwo}%

\theoremstyle{thmstylethree}%

\begin{document}

\title[Article Title]{KFFocus: Highlighting Keyframes for Enhanced Video Understanding}

\author[1]{\fnm{Ming} \sur{Nie}}
\author[2]{\fnm{Chunwei} \sur{Wang}}
\author[2]{\fnm{Hang} \sur{Xu}}
\author[1]{\fnm{Li} \sur{Zhang}}

\affil[1]{School of Data Science, Fudan University}

\affil[2]{Huawei}

\abstract{Recently, with the emergence of large language models, multimodal LLMs have demonstrated exceptional capabilities in image and video modalities.
Despite advancements in video comprehension, the substantial computational demands of long video sequences lead current video LLMs (Vid-LLMs) to employ compression strategies at both the inter-frame level (\eg, uniform sampling of video frames) and intra-frame level (\eg, condensing all visual tokens of each frame into a limited number).
However, this approach often neglects the uneven temporal distribution of critical information across frames, risking the omission of keyframes that contain essential temporal and semantic details.
To tackle these challenges, we propose KFFocus, a method designed to efficiently compress video tokens and emphasize the informative context present within video frames.
We substitute uniform sampling with a refined approach inspired by classic video compression principles to identify and capture keyframes based on their temporal redundancy.
By assigning varying condensation ratios to frames based on their contextual relevance, KFFocus efficiently reduces token redundancy while preserving informative content details.
Additionally, we introduce a spatiotemporal modeling module that encodes both the temporal relationships between video frames and the spatial structure within each frame, thus providing Vid-LLMs with a nuanced understanding of spatial-temporal dynamics.
Extensive experiments on widely recognized video understanding benchmarks, especially long video scenarios, demonstrate that KFFocus significantly outperforms existing methods, achieving substantial computational efficiency and enhanced accuracy.}


\maketitle

\section{Introduction}
Large language models (LLMs) have gained significant attention for their impressive text understanding capabilities.
Leveraging the strengths of LLMs, video large language models (Vid-LLMs)~\cite{li2023videochat, lin2023videollava, zhang2023videollama} adapt these techniques to videos, thereby extending their reasoning and interactive abilities to the video data.
Through training on video-level tasks such as captioning and question-answering, Vid-LLMs establish promising video-language correspondence and enhance their video understanding abilities.

\begin{figure}[t]
    \begin{center}
        \includegraphics[width=1.0\linewidth]{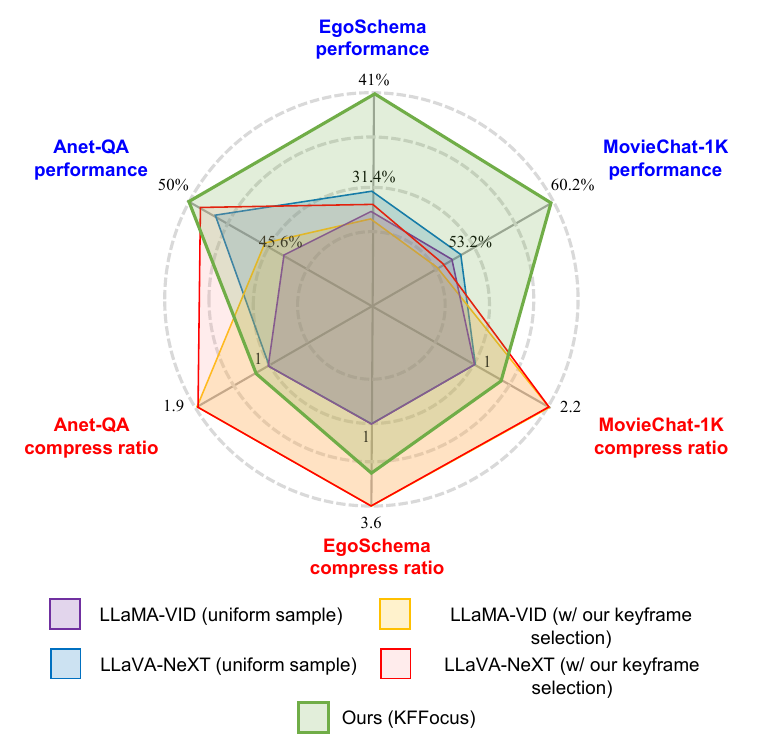}
    \end{center}
    \caption{Empirical comparison of uniform sampling and keyframe selection.
    The performance and visual token cost (indicated by token compression ratio) on benchmarks are highlighted in \textcolor{blue}{blue} and \textcolor{red}{red} color, respectively.
    Compared with uniform sampling, methods employing our proposed keyframe selection notably reduce computational costs while maintaining comparable performance.
    Moreover, our KFFocus achieves leading results, demonstrating the rationality behind our approach for keyframe-centered efficient video token compression.}
    \label{fig-intro1}
\end{figure}

\begin{figure*}[t]
    \begin{center}
        \includegraphics[width=1.0\linewidth]{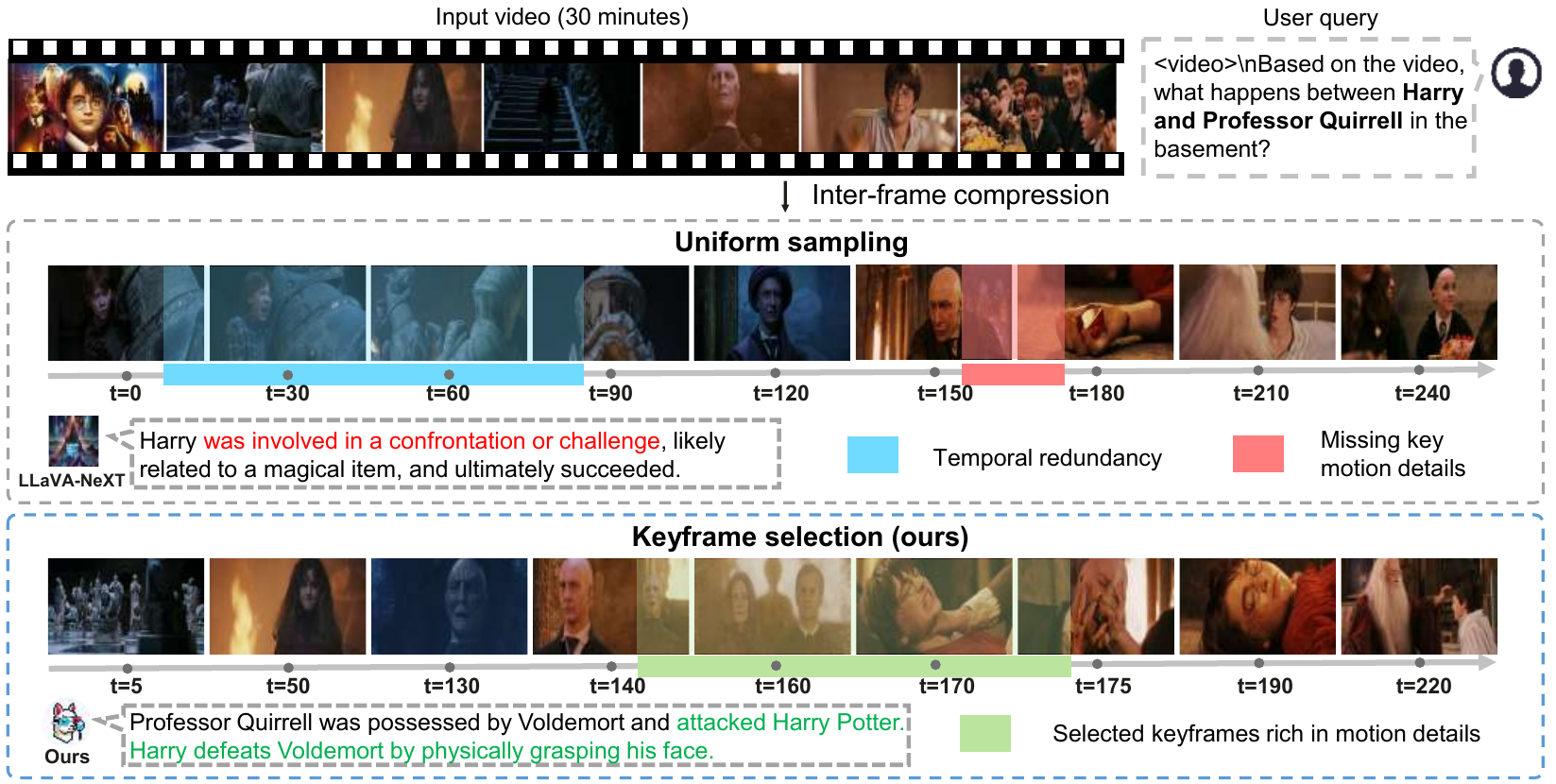}
    \end{center}
    \caption{Illustration of uniform sampling on timestamp versus keyframe selection. Uniform sampling on long video sequences risks both frame redundancy and missing motion details, while our keyframe selection reduces redundancy and more effectively captures motion dynamics.}
    \label{fig-intro2}
\end{figure*}

While Vid-LLMs have shown impressive performance in video comprehension, they still face a significant challenge.
To embed video features into LLMs under computing resource constraints, the input videos typically undergo both inter-frame and intra-frame compression.
Vid-LLMs often uniformly sample video frames along the temporal dimension and compress the spatial tokens of each sampled frame to a fixed number.
This process operates on the assumption that video information is evenly distributed.
However, such an assumption contradicts reality: (i) redundancy exists among successive video frames and the motion pattern of each video clip persists for varying durations, resulting in an uneven distribution of keyframes, and (ii) the significance and relevance of various video frames exhibit inconsistent variation in relation to the specific query.
For instance, given a query, the related video clip may concentrate within only a few seconds, making those frames highly informative.
In such case, uniform inter-frame compression risks omitting these keyframes and losing essential temporal details, while uniform token condensation may weaken the important semantic and spatial context within frames.
In contrast, selecting essential video frames that reflects the temporal dynamics of the video and allocating more visual tokens to represent contextually relevant frames presents a more promising approach.

To address these issues, we propose KFFocus (Key Frame Focus) to efficiently compress video tokens and highlight informative details within video frames.
We introduce a novel hybrid token compression strategy that replaces conventional uniform temporal sampling and even token condensation.
Specifically, we develop an inter-frame compression method that extracts keyframes based on their temporal dependencies.
Inspired by traditional video compression techniques, where adjacent video frames are typically encoded into a single keyframe, this method analyzes the visual differences between each frame and its predecessors to identify the most informative keyframes.
The sampled keyframes are temporally independent, capturing essential motion content of the input video.
Extensive experimental results, as shown in Figure~\ref{fig-intro1}, reveal that selected keyframes notably reduce computational costs while maintaining comparable performance compared to uniform sampling.
These advantages are particularly pronounced on long video benchmarks, underscoring the potential of keyframe selection.
In addition, to address the limitation of uniform token condensation, we introduce a dynamic token condensation module.
Considering that frames relevant to the user question deserve more tokens to represent their significance, we assign varying token condensation ratios based on each frame’s semantic importance.
This approach allows for a more focused and effective representation of keyframes.
The visualizations in Figure~\ref{fig-intro2} highlight our method's ability to capture frames that are rich in temporal dynamics and semantic content.

To enable LLMs to fully capture the inter-frame and intra-frame relationships of visual tokens within our hybrid token compression framework, we further propose a spatial-temporal modeling module.
Unlike conventional methods that implicitly model temporal relationships among input visual embeddings based on their sequential positions, this module explicitly encodes both the relative temporal positions of video frames and the spatial structures within each frame's patches.
By informing the model of these spatial and temporal structures, we enhance our hybrid token compression strategy, enabling more nuanced video comprehension and ultimately achieving more accurate and meaningful video analysis.
As illustrated in Figure~\ref{fig-intro1}, our method delivers superior performance with reduced visual token usage, highlighting its effectiveness.
These benefits are especially evident on long video benchmarks, where our approach significantly outperforms others, validating the rationale behind our method for efficient token compression in long video sequences.

The contributions of this paper are summarized as follows:
\textbf{(i)}
We propose a hybrid token compression strategy that identifies and selects keyframes based on temporal redundancy.
This strategy also assigns varying condensation ratios to frame tokens based on their contextual relevance towards specific user questions, enhancing the representation of critical information.
\textbf{(ii)}
We introduce a novel spatiotemporal modeling module to explicitly encode the temporal relationships and spatial structure of video content.
By enhancing LLMs' awareness of the relative positions of video frames and the spatial layout of frame patches, this approach improves the model’s ability to capture complex spatial-temporal dynamics, leading to more accurate video comprehension.
\textbf{(iii)}
We conduct extensive experiments to evaluate the effectiveness of KFFocus across various widely accepted video understanding benchmarks.
Our method outperforms current state-of-the-art models, particularly on long video benchmarks, demonstrating the efficiency of our hybrid token compression strategy and spatial-temporal modeling module.

\section{Related works}
\noindent\textbf{Vision large language models.}
Researchers have made substantial progress in enabling Large Language Models (LLMs) to understand visual information.
BLIP-2~\cite{li2023blip} introduces Q-Former, a lightweight Querying Transformer that aligns vision-language representation, while MiniGPT-4~\cite{zhu2023minigpt} links detailed image descriptions with advanced LLMs, significantly boosting multimodal capabilities.
By leveraging diverse multimodal instruction-following data, LLaVA~\cite{liu2024visual} and LLaVA-NeXT~\cite{liu2024llavanext} have demonstrated impressive multimodal conversational abilities.
Recent models, such as Kosmos-2~\cite{peng2023kosmos} and VisionLLM~\cite{wang2024visionllm}, have further refined image comprehension skills, including referring and grounding, thereby enhancing the ability to describe complex image details.

\noindent\textbf{Video large language models.}
The exploration of LLMs has progressed from static images to dynamic video data, leading to the rise of Vid-LLMs.
VideoChat~\cite{li2023videochat} combines fine-tuning with LLM-based video agents and is refined using a specially crafted video-centric instruction dataset.
Video-ChatGPT~\cite{maaz2023videochatgpt} introduces a semi-automated, human-assisted annotation framework to generate high-quality instructional data for video tasks.
Video-LLaMA~\cite{zhang2023videollama}, trained independently on vision-language and audio-language branches with consistent visual processing, has shown impressive abilities in understanding both visual and auditory content.
Video-LLaVA~\cite{lin2023videollava} achieves unified visual representation through alignment before projection and performs joint training on images and videos.
Together, these advancements have equipped Vid-LLMs with robust capabilities in downstream tasks such as text-video retrieval and video captioning.
However, these models still struggle to fully comprehend long videos in both fine-grained temporal and spatial aspects.
To address this challenge, we introduce a model with robust capabilities for detailed spatiotemporal video understanding.

\noindent\textbf{Keyframe selection for video understanding.}
Vid-LLMs often employ inter-frame and intra-frame compression to mitigate computing resource constraints.
Most existing methods~\cite{lin2023videollava,wang2023internvid, zhang2023videollama, wang2024internvideo2} rely on uniform sampling or random frame selection to fix the number of input frames during both training and testing.
Additionally, some approaches~\cite{li2023videochat,ataallah2024minigpt4video} utilize token condensation to compress the spatial tokens of each selected frame, leveraging visual adapters like average pooling~\cite{liu2024visual} or Q-former~\cite{li2023blip}.
However, these techniques are often inadequate for capturing the most informative frames and preserving crucial visual tokens, as they overlook the temporal redundancies and uneven distribution of information typical in video data.
Meanwhile, some approaches integrate frame selection strategy to enhance the quality of sampled frames.
VideoChatGPT~\cite{maaz2023videochatgpt} uses Katna~\cite{katna}, a machine learning-based keyframe selection method that is highly sensitive to parameter tuning.
Video-LaVIT~\cite{jin2024video} combines keyframes with motion vectors for video tokenization, while KeyVideoLLM~\cite{liang2024keyvideollm} employs a CLIP-based technique to identify keyframes, though it incurs substantial additional computional costs.
In this work, we propose a hybrid token compression strategy that simultaneously addresses video token redundancy from both inter-frame and intra-frame perspective, while ensuring the retention of tokens rich in temporal and semantic content.

\section{Method}
\begin{figure*}[t]
    \begin{center}
        \includegraphics[width=1.0\linewidth]{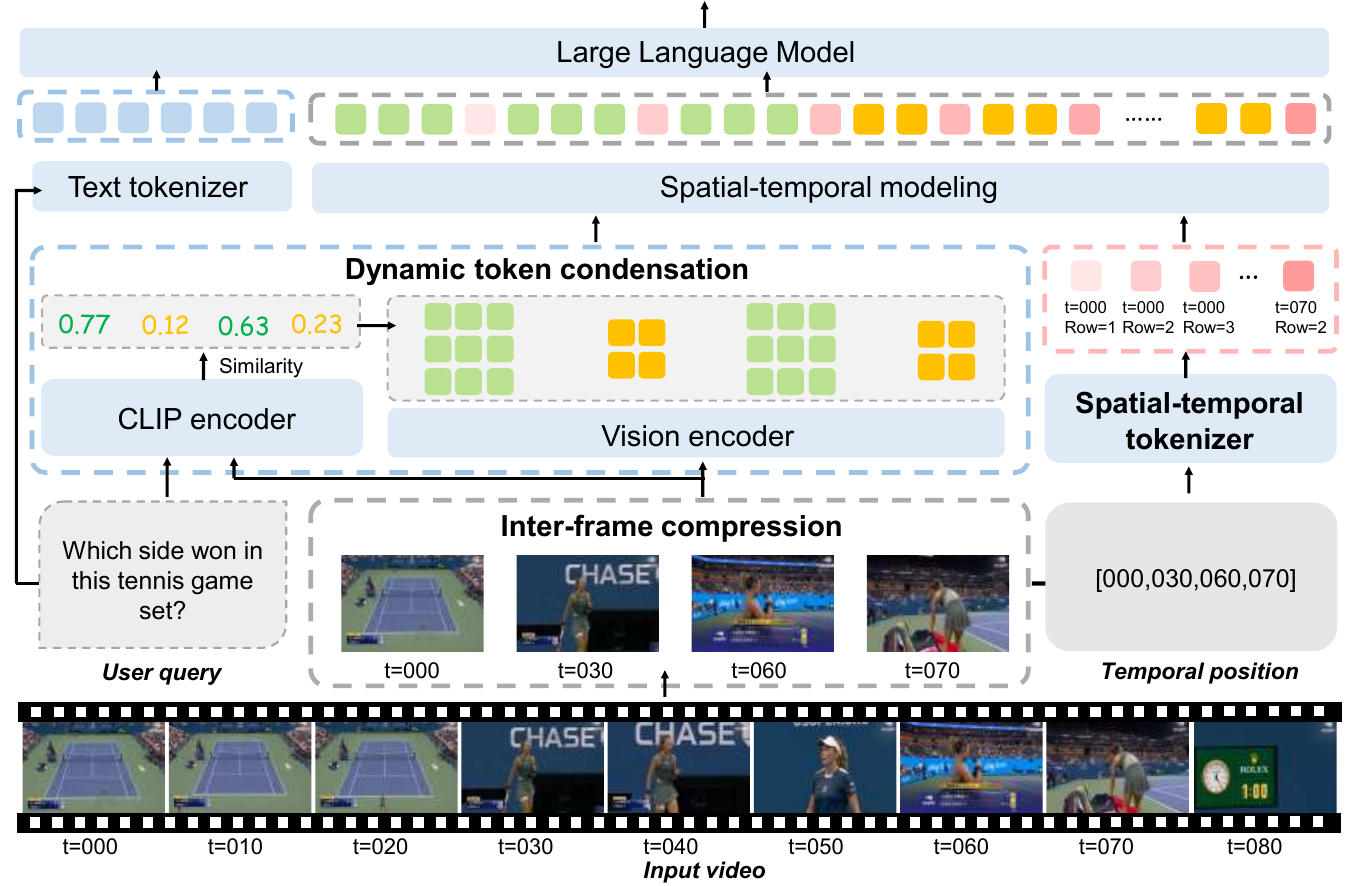}
    \end{center}
    \caption{The framework of KFFocus. We begin by sampling keyframes and emphasizing contextual details through our inter-frame compression and dynamic token condensation component. Our spatial-temporal modeling module then organizes the hybrid visual embeddings to explicitly inform LLMs the spatiotemporal structure of input tokens.}
    \label{fig-method}
\end{figure*}
In this section, we introduce KFFocus, a large multimodal model that can efficiently compress video frames and focus on informative video tokens.
KFFocus aims to enhance video compression efficiency by highlighting the essential spatiotemporal information of keyframes while reducing the redundant processing of non-keyframes.
In pursuit of this goal, we propose two core designs: a hybrid token compression strategy (Section~\ref{hybrid}) which allows for a hierarchical video tokenization process, and a spatial-temporal modeling module (Section~\ref{stmodel}) to organize video tokens for LLMs.
Coped with these two main components, KFFocus can be optimized through a unified autoregressive training paradigm (Section~\ref{train}).

\subsection{Hybrid Token Compression}\label{hybrid}
\noindent\textbf{Preliminary.}
Contemporary Vid-LLMs typically feature a modular architecture, which includes a visual encoder $E_{V}$, a series of visual adapters $Q$, and a large language model $L$.
For a given video $V = \{V(t) \in \mathbb{R}^{H \times W \times 3} | t=0,...,T \}$ that consists of $T$ frames, along with its associated question $q$, current Vid-LLMs generally perform inter- and intra-frame compression on the original video $V$.
Due to significant redundancy in temporal dimension of videos, only a limited number of frames are typically selected as the actual input for the Vid-LLMs.
This process involves evenly sampling the video frames along the temporal dimension, resulting in a subset of video frames $V'$:

\begin{equation}
V' = \{V(Mt) \in \mathbb{R}^{H \times W \times 3} | t=0,...,T \},
\end{equation}
where $M$ is the fixed sampling interval.
Subsequently, the visual encoder processes the downsampled frames $V'$ and encodes them into a series of visual tokens denoted as $z = E_{V}(V')$.
These visual tokens are then transformed to align with the embedding space of the language model through the visual adapter $Q$, and concatenated with text tokens as the input for the language model decoder.

Although this paradigm is well developed in the field of video understanding, it encounters significant limitations.
First, uniform sampling can result in a considerable amount of redundancies in the input for LLMs.
Second, due to the fact that important contexts are not evenly distributed temporally, uniform sampling does not ensure that keyframes will be properly retained.
To address these shortcomings, we revisit the inherent properties of video data and develop an inter-frame compression strategy to extract keyframes that reflect the temporal and spatial characteristics of the video.
This strategy is introduced as follows.

\noindent\textbf{Inter-frame compression.}
Considering the temporal redundancy and motion relation in adjacent video frames, we draw inspiration from the traditional video compression protocol~\cite{sikora1997mpeg}, where I-frames serve as fundamental implementation.
I-frames in the video compression process are complete, self-contained images that can be independently decoded, retaining important visual information in the video.
We extend it to video understanding tasks by replacing the original uniform sampling with I-frames, which are selected by analyzing the visual differences between each frame and its preceding one.
Further details on the extraction of I-frames are provided in the appendix.

After extracting the I-frames, which are represented as $I$, we enhance the compression strategy by incorporating compensation frames to ensure effective coverage of content changes in the video.
Compensation frames are predictive frames that rely on previous I-frames for video reconstruction.
Based on extracted I-frames, we compute the time interval $T_k$ between adjacent I-frames to determine the amount of compensation frames to be sampled:

\begin{equation}
    n = \lfloor \frac{T_k}{\delta T} \rfloor,
\end{equation}
where $\delta$ is the coverage ratio of the whole video.
We insert the additional frames into $I$ to serve as temporal compensations.
Subsequently, $I$ is fed into visual encoder and our post-processing.

\noindent\textbf{Dynamic token condensation.}
Long videos require LLMs to process significantly more visual tokens, which constitutes a major portion of the computation.
To address this issue, the encoded visual tokens of each frame $z \in \mathbb{R}^{N \times C}$ are condensed to a lower number $N'$ using spatial pooling: $z' = pool(z)$.
The condensed token number $N'$ equals $N/d^2$, where $d$ represents the spatial down-sample ratio.

Considering that the sampled video frames are temporally independent yet may not be relevant to the user query, retaining a constant down-sample ratio is computationally inefficient.
To address this, we propose a dynamic token condensation strategy.
As illustrated in Figure~\ref{fig-method}, we assign varying token condensation ratios based on each frame’s contextual significance.
Inspired by~\cite{liang2024keyvideollm}, we compute the embedding correlation between the sampled keyframes and the input user query using CLIP~\cite{radford2021learning}.
The top $\alpha$ portion of keyframes are selected based on semantic similarity, where $\alpha$ represents the contextual focusing ratio.
For frames with higher correlation, we assign a lower condensation ratio $d_1$ to enhance our Vid-LLM's focus on these frames.
For the remaining frames, a higher condensation ratio $d_2$ is applied to balance efficient temporal coverage and manage computational resources effectively.

\subsection{Spatial-Temporal Modeling for Video Tokens}\label{stmodel}
\noindent\textbf{Discretized tokenizer.}
While LLMs can implicitly capture temporal relationships among input visual embeddings based on their sequential positions, they encounter difficulties when the visual tokens are not evenly distributed along the timeline.
To address this challenge, we propose a discretized tokenizer that encodes the relative positions of frames into a set of discretized temporal tokens $\mathcal{E} \in \mathbb{R}^{\mathcal{T} \times C}$, where $\mathcal{T}$ represents the discrete temporal space.
For keyframes sampled at multiple frequencies $I_k$, the corresponding temporal token is $\epsilon_{k}$, where $\epsilon_{k} = \mathcal{E}[\lfloor \mathcal{T} * k/T \rfloor]$.
These temporal embeddings are then incorporated with the visual tokens to model the spatial-temporal relationship in an explicit manner.

\noindent\textbf{Spatial-temporal modeling.}
Since the video tokens is hybrid across different frames, it is necessary to inform LLMs of the spatial organization of video tokens, including both inter-frames and intra-frames.
Inspired by~\cite{xu2024llavauhd}, we develop a spatial-temporal modeling schema to inform the relative temporal positions of video frames and spatial relation of frame patches to LLMs.
As is illustrated in Figure~\ref{fig-method}, specifically, in each frame $I_k$, we utilize $\epsilon^{intra}_{k} = g_{intra}(\epsilon_{k})$ to separate the slice representations in frame tokens:

\begin{equation}
    z_{row,k} = Cat(z_{row,k}, \epsilon^{intra}_{k}),
\end{equation}
where $g(\cdot)$ is a mlp projector and $z_{row,k} \in \mathbb{R}^{w \times C}$ indicates the $row$-th slice of visual tokens in frame $I_k$.
Similarly, we leverage $\epsilon^{inter}_{k} = g_{inter}(\epsilon_{k})$ to separate video embeddings between adjacent frames:

\begin{equation}
    z_k = Cat(z_k, \epsilon^{inter}_{k}).
\end{equation}
Then the visual tokens are flatten and concatenated with text tokens as the input for LLMs.
In our experiments, we find that this simple modeling can effectively leverage the hybrid visual tokens to yield good performance.

\subsection{Training Strategy}\label{train}
Considering training efficiency, our training procedure is structured into two distinct phases in this work: (i) pre-training for modality alignment and (ii) fine-tuning for enhancing temporal video understanding.
We now provide detailed elaborations on the training strategies and datasets employed for each of these stages.

\noindent\textbf{Modality alignment.}
In the pre-training stage, our primary focus is to optimize the visual adapters and temporal tokenizer, while the visual encoder and language model are frozen to ensure that the visual features align effectively with the language space.
Following the approach used in LLaMA-VID~\cite{li2023llamavid}, we utilize the image-text LCS-558K dataset from LLaVA~\cite{liu2024visual}, along with 232K video-caption samples from the WebVid 2.5M~\cite{bain2021frozen}.

\begin{table*}[t]
\renewcommand\arraystretch{1.5}
\begin{center}
\resizebox{1.0\linewidth}{!}{
\begin{tabular}{ccc|cc|cc|cc|ccccc}
\midrule[1.0pt]
& & & \multicolumn{2}{c|}{MSVD-QA} & \multicolumn{2}{c|}{MSRVTT-QA} & \multicolumn{2}{c|}{ActivityNet-QA} & \multicolumn{5}{c}{Video-based generative performance} \\
\multirow{-2}{*}{Method} & \multirow{-2}{*}{LLM} & \multirow{-2}{*}{LoRA} & Acc & Score & Acc & Score & Acc & Score & Correctness & Detail & Context & Temporal & Consistency \\
\midrule[1.0pt]
FrozenBiLM~\cite{yang2022zero} & DeBERTa-V2 & \xmark & 32.2 & - & 16.8 & - & 24.7 & - & - & - & - & - & - \\
Video-LLaMA~\cite{zhang2023videollama} & Vicuna-7B & \xmark & 51.6 & 2.5 & 29.6 & 1.8 & 12.4 & 1.1 & 1.96 & 2.18 & 2.16 & 1.82 & 1.79 \\
LLaMA-Adapter~\cite{zhang2023llama} & LLaMA-7B & \xmark & 54.9 & 3.1 & 43.8 & 2.7 & 34.2 & 2.7 & 2.03 & 2.32 & 2.30 & 1.98 & 2.15 \\
VideoChat~\cite{li2023videochat} & Vicuna-7B & \xmark & 56.3 & 2.8 & 45.0 & 2.5 & 26.5 & 2.2 & 2.23 & 2.50 & 2.53 & 1.94 & 2.24 \\
Video-ChatGPT~\cite{maaz2023videochatgpt} & Vicuna-7B & \xmark & 64.9 & 3.3 & 49.3 & 2.8 & 35.2 & 2.7 & 2.40 & 2.52 & 2.62 & 1.98 & 2.37 \\
LLaMA-VID~\cite{li2023llamavid} & Vicuna-7B & \xmark & 69.7 & 3.7 & 57.7 & 3.2 & 47.4 & 3.3 & 2.96 & 3.00 & 3.53 & 2.46 & 2.51 \\
\midrule
LLaMA-VID & Vicuna-7B & \cmark & 69.2 & 3.4 & 57.1 & 2.9 & 45.6 & 3.3 & 2.87 & 2.89 & 3.28 & 2.13 & 2.47 \\
LLaMA-VID\dag & Vicuna-7B & \cmark & 69.4 & 3.3 & 56.8 & 2.8 & 46.1 & 3.4 & 2.89 & 2.85 & 3.17 & 2.16 & 2.43 \\
\midrule
LLaVA-NeXT~\cite{liu2024llavanext} & Llama3-8B & \cmark & 69.8 & 3.5 & 58.6 & 2.9 & 47.4 & 3.4 & 2.91 & 2.85 & 3.15 & 2.28 & 2.45 \\
LLaVA-NeXT\dag & Llama3-8B & \cmark & 69.7 & 3.4 & 58.3 & 2.9 & 47.8 & 3.4 & 2.90 & 2.87 & 3.10 & 2.19 & 2.41 \\
\midrule
Ours & Llama3-8B & \cmark & 70.5 & 3.6 & 59.3 & 3.1 & 49.8 & 3.6 & 3.01 & 3.02 & 3.59 & 2.55 & 2.57 \\
Ours & Qwen2-7B & \cmark & \textbf{70.8} & \textbf{3.7} & \textbf{60.0} & \textbf{3.3} & \textbf{51.2} & \textbf{3.7} & \textbf{3.13} & \textbf{3.09} & \textbf{3.62} & \textbf{2.74} & \textbf{2.83} \\
\midrule[1.0pt]
\end{tabular}}
\end{center}
\caption{Comparison with existing methods on short video understanding benchmarks. \dag~indicates inter-frame compression is performed during inference. Our method achieve advanced performance compared to state-of-the-art models.}
\label{tab-coarse}
\end{table*}

\noindent\textbf{Temporal enhancement.}
After the pre-training stage, the Vid-LLM gains proficiency in processing visual information.
During the fine-tuning stage, we focus on enhancing the model's ability to comprehend sequences of video frames, thereby improving its temporal video understanding.
In line with the practices from Video-ChatGPT~\cite{maaz2023videochatgpt} and LLaMA-VID~\cite{li2023llamavid}, we employ the instruction datasets based on three sources: 40K text conversations, 625K visual question-answer pairs (both single-turn and multi-turn), and 98K video QA pairs.
These instructions are augmented with different formats for diversity.

Furthermore, we utilize the InternVid-10M-FLT dataset~\cite{wang2023internvid}, which is specifically crafted for training with a focus on temporal-awareness~\cite{huang2023vtimellm}.
We find that training on these temporally sensitive tasks help our spatial-temporal modeling component better learn and capture the temporal relationships among video frame tokens.
Throughout this stage, we continue to train the visual adapters and temporal tokenizer.
Additionally, the LLM is further trained using LoRA~\cite{hu2021lora} to refine its capabilities.

\subsection{Implementation Details}
We provide more implementation details of our KFFocus in this section.

\noindent\textbf{I-frames extraction.}
We utilize FFmpeg to identify I-frames, which are encoded as key frames within the Group of Pictures (GoP) structure in video streams.
FFmpeg efficiently analyzes the frame types, distinguishing between I-, P-, and B-frames, to extract I-frames that serve as independent reference points in compressed videos.
To ensure high-quality I-frame selection, we leverage Structural Similarity Index (SSIM) and Peak Signal-to-Noise Ratio (PSNR)~\cite{hore2010image-psnr}, which are widely used metrics for assessing image quality.
These metrics guide FFmpeg in adjusting I-frame compression quality, ensuring that key frames retain essential visual details while minimizing compression artifacts.
By incorporating SSIM and PSNR, we enhance the robustness of I-frame extraction, preserving critical structural and perceptual information necessary for downstream video understanding tasks.

\noindent\textbf{Hybrid token condensation.}
For keyframe selection, we adopt the MPEG-4~\cite{sikora1997mpeg} compression technique to identify keyframes.
The coverage ratio $\alpha$ is set to 5\%, which determines the proportion of compensation frames to be sampled.
We utilize a pre-trained CLIP model to extract textual embeddings and compute their similarity with visual embeddings.
Based on these similarity scores, we rank all frames and select the top $\alpha$ with the highest scores, assigning more visual tokens to these contextually relevant frames.
As a counterpart, for uniform sampling, we compress the video by setting the fps rate to 1 and condense each frame into 16 tokens.

\noindent\textbf{Visual encoder.}
During training and inference, keyframes are resized to a resolution of 224×224 for input.
We utilize CLIP-ViT-L-14~\cite{radford2021learning} as the frozen visual encoder, with a patch size of 14, resulting in an original token count $N=256$ per frame.
Contextually relevant frames are condensed to 64 tokens using $d_1$, while the remaining frames are condensed to 16 tokens using $d_2$, both through spatial average pooling.
For visual adapters, we adopt the approach from LLaVA~\cite{liu2024visual}, training a projector layer to align visual features with the pre-trained LLM's word embedding space.
These visual adapters are trained during both the pre-training and fine-tuning phases.

\noindent\textbf{Training details.}
During pre-training, we update the linear layer that projects video features to the LLMs’ input space, as well as the spatial-temporal modeling module, which includes a temporal tokenizer and two projector layers, while keeping the rest of the architecture frozen.
The model is pre-trained on the pre-training dataset for one epoch with a learning rate of $1e-3$ and an overall batch size of 32.
In the fine-tuning stage, we utilize LoRA to adapt the LLM's parameters.
The LoRA configuration is set with $r = 64$ and $\alpha = 128$.
Fine-tuning is conducted for one epoch with a learning rate of $2e-4$ and an overall batch size of 8.

\section{Experiments}
\begin{table}[t]
\renewcommand\arraystretch{1.5}
\begin{center}
\resizebox{1.0\linewidth}{!}{
\begin{tabular}{ccc|cc|cc}
\midrule[1.0pt]
& & & \multicolumn{2}{c|}{Global mode} & \multicolumn{2}{c}{Breakpoint mode} \\
\multirow{-2}{*}{Method} & \multirow{-2}{*}{LLM} & \multirow{-2}{*}{LoRA} & Acc & Score & Acc & Score \\
\midrule[1.0pt]
VideoChat~\cite{li2023videochat} & Vicuna-7B & \xmark & 57.8 & 3.00 & 46.1 & 2.29 \\
VideoLLaMA~\cite{zhang2023videollama} & Vicuna-7B & \xmark & 51.7 & 2.67 & 39.1 & 2.04\\
Video-ChatGPT~\cite{maaz2023videochatgpt} & Vicuna-7B & \xmark & 47.6 & 2.55 & 48.0 & 2.45 \\
MovieChat~\cite{song2024moviechat} & Vicuna-7B & \xmark & 62.3 & 3.23 & 48.3 & 2.57 \\
\midrule
LLaMA-VID~\cite{li2023llamavid} & Vicuna-7B & \cmark & 52.8 & 2.85 & 41.5 & 2.21 \\
LLaMA-VID\dag & Vicuna-7B & \cmark & 52.1 & 2.79 & 41.1 & 2.19 \\
\midrule
LLaVA-NeXT~\cite{liu2024llavanext} & Llama3-8B & \cmark & 53.2 & 2.86 & 41.3 & 2.18 \\
LLaVA-NeXT\dag & Llama3-8B & \cmark & 52.3 & 2.80 & 41.0 & 2.17 \\
\midrule
Ours & Llama3-8B & \cmark & 60.2 & 3.20 & 47.6 & 2.49 \\
Ours & Qwen2-7B & \cmark & \textbf{62.7} & \textbf{3.32} & \textbf{49.4} & \textbf{2.59} \\
\midrule[1.0pt]
\end{tabular}}
\end{center}
\caption{Comparison with existing methods on MovieChat-1K. \dag~indicates inter-frame compression is performed during inference.}
\label{tab-moviechat}
\end{table}
\begin{table}[t]
\renewcommand\arraystretch{1.5}
\begin{center}
\resizebox{0.8\linewidth}{!}{
\begin{tabular}{ccc|c}
\midrule[1.0pt]
Method & LLM & LoRA & Acc \\
\midrule[1.0pt]
FrozenBiLM~\cite{frozenbilm} & DeBERTa~\cite{he2020deberta} & \xmark & 26.9 \\
VIOLET~\cite{fu2021violet} & - & \xmark & 19.9 \\
InternVideo~\cite{wang2022internvideo} & - & \xmark & 32.1 \\
LLoVi-7B~\cite{zhang2023llovi} & Llama2-7B~\cite{touvron2023llama2} & - & 34.0 \\
Vamos~\cite{wang2023vamos} & GPT-4 & - & 36.7 \\
\midrule
Ours & Llama3-8B & \cmark & 55.1 \\
Ours & Qwen2-7B & \cmark & \textbf{58.2} \\
\midrule[1.0pt]
\end{tabular}}
\end{center}
\caption{Comparison with existing methods on EgoSchema.}
\label{tab-egoschema}
\end{table}
\begin{table*}[h]
\renewcommand\arraystretch{1.5}
\begin{center}
\resizebox{0.8\linewidth}{!}{
\begin{tabular}{c|cc|c|ccc|ccc}
\midrule[1.0pt]
& \multicolumn{2}{c|}{Hybrid token compression} & & \multicolumn{3}{c|}{ActivityNet-QA} & \multicolumn{3}{c}{MovieChat-1K} \\
\multirow{-2}{*}{Row} & Inter-frame & Dynamic token & \multirow{-2}{*}{ST modeling} & Token \# & Acc & Score & Token \# & Acc & Score \\
\midrule[1.0pt]
I & \xmark & \xmark & \xmark & $\sim$3.6k & 47.0 & 3.3 & $\sim$8.4k & 53.4 & 2.8 \\
II & \cmark & \xmark & \xmark & $\sim$1.9k & 47.2 & 3.3 & $\sim$3.8k & 52.9 & 2.8 \\
\midrule
III & \xmark & \xmark & \xmark & $\sim$2.8k & 44.6 & 3.1 & $\sim$5.1k & 50.2 & 2.7 \\
IV & \cmark & \xmark & \xmark & $\sim$2.8k & 46.8 & 3.3 & $\sim$5.1k & 52.4 & 2.7 \\
V & \cmark & \cmark & \xmark & $\sim$2.8k & 48.3 & 3.4 & $\sim$5.1k & 55.4 & 2.9 \\
\midrule
\rowcolor[gray]{0.9}VI & \cmark & \cmark & \cmark & $\sim$3.0k & 49.8 & 3.6 & $\sim$5.5k & 60.2 & 3.2 \\
\midrule[1.0pt]
\end{tabular}}
\end{center}
\caption{Ablation on token compression strategies. \# indicates the average visual token number on the corresponding benchmark.}
\label{tab-component}
\end{table*}
\subsection{Experiment Setup}
In our experiments, we implement Llama3-8B~\cite{touvron2023llama} as our foundation LLM and Qwen2-7B~\cite{yang2024qwen2} as a more powerful LLM to demonstrate the effectiveness and extensibility of our method.
We adjust the resolution of input videos to 224 $\times$ 224, and employ CLIP-ViT-L-14~\cite{radford2021learning} for the frozen visual encoder.
The patch size is set to 14 and the original token number $N$ is 256 accordingly.
We assign the lower condensation ratio $d_1$ to 2 and the higher condensation ratio $d_2$ to 4, which results in 64 tokens for the contextual relevant frames and 16 tokens for the rest.
For visual adapters, we follow LLaVA~\cite{liu2024visual} to train a projector layer, aligning visual features with the pre-trained LLM word embedding.
The size of temporal token space $\mathcal{T}$ is set as 1000.
The AdamW~\cite{loshchilov2017decoupled} optimizer is applied with a cosine learning rate and decay and a warm-up period.
We train our Vid-LLM for two stages.
During the initial pre-training stage, the learning rate is set to $1e-3$.
For the subsequent fine-tuning stages, the learning rate is adjusted to $2e-4$.
Additionally, the LoRA parameters are configured with $r = 64$ and $alpha = 128$.
For a comprehensive comparison, we also implement several state-of-the-art frameworks, including LLaMA-VID~\cite{li2023llamavid} and LLaVA-NeXT~\cite{liu2024llavanext}.
These models are trained on the same datasets and fine-tuned with LoRA to ensure a fair evaluation.
All experiments are conducted on 8 V100 GPUs.

\subsection{Evaluation Details}
To evaluate the understanding capabilities in video-based benchmarks, we consider six common benchmarks: MSVD-QA~\cite{xu2017video}, MSRVTT-QA~\cite{xu2016msr}, ActivityNet-QA~\cite{yu2019activitynet} and video-based generative benchmark~\cite{maaz2023videochatgpt} for short-video scenarios, along with MovieChat-1K~\cite{song2024moviechat} and EgoSchema~\cite{mangalam2023egoschema} for long-video scenarios.

For the video-based generative benchmark introduced by VideoChatGPT~\cite{maaz2023videochatgpt}, the evaluation pipeline utilizes the GPT-3.5 model.
This pipeline assesses the model's capabilities across five key aspects: information correctness, detail orientation, contextual understanding, temporal understanding, and consistency, assigning a relative score to each generated prediction on a scale of 1 to 5.

For EgoSchema, the benchmark consists of multiple-choice questions, each accompanied by five or more possible answers.
The model is required to identify and select the most appropriate answer for each question.
To facilitate this evaluation, we concatenate each specific answer with the corresponding question, querying the LLM about the correctness of the given answer.
We then extract the logit value of the hidden state associated with the ``Yes'' or ``yes'' response to determine the model's confidence in the correctness of the answer, ultimately selecting the one with the highest confidence as the final prediction.

For the remaining benchmarks, the evaluation follows a single-turn QA process.
Additionally, the GPT-3.5 model is employed to assess the correctness and quality of the generated answers, with performance reported in terms of accuracy and relative scores on a scale of 1 to 5.

\subsection{Main Results}
\noindent\textbf{Results on short video benchmarks.}
We begin by conducting a comparative evaluation of our method against various state-of-the-art approaches across three zero-shot video-QA benchmarks: MSVD-QA~\cite{xu2017video}, MSRVTT-QA~\cite{xu2016msr}, and ActivityNet-QA~\cite{yu2019activitynet}.
Additionally, we assess performance on the video-based generative benchmark~\cite{maaz2023videochatgpt}.
These benchmarks consist of short videos lasting seconds to minutes.
The evaluations are performed in a zero-shot manner, utilizing GPT-assisted evaluation to assess the model's capabilities.
As illustrated in Table~\ref{tab-coarse}, our KFFocus method achieves superior performance across various short video benchmarks compared to other approaches.
Additionally, we investigate the quality of inter-frame compression by conducting evaluations where models are trained with normal uniform sampling and tested using only I-frames as input (noted as \dag).
Notably, experimental evidence prove that inference using I-frames also yields comparable results, confirming the informative value of inter-frame compression.
This effect is especially pronounced on the ActivityNet-QA benchmark, where I-frames inference results in a performance improvement (+0.5\% for LLaMA-VID and +0.4\% for LLaVA-NeXT).

\noindent\textbf{Results on long video benchmarks.}
To further investigate the effectiveness of the proposed method in more challenging scenarios, we provide evaluations on a diverse set of long video benchmarks, including MovieChat-1K~\cite{song2024moviechat}, EgoSchema~\cite{mangalam2023egoschema}, as well as MLVU~\cite{zhou2024mlvu}, MVBench~\cite{li2024mvbench}, VideoMME~\cite{fu2024videomme} and LVBench~\cite{wang2024lvbench}.
The results for the latter four benchmarks are provided in the \textbf{\textit{appendix}}.
As shown in Table~\ref{tab-moviechat}, we evaluate our method on MovieChat-1K.
The results demonstrate that although our method is not specifically trained on long video datasets and is evaluated in a zero-shot setting on MovieChat-1K benchmark (in contrast, MovieChat~\cite{song2024moviechat} has undergone targeted training for long videos), it still achieves competitive performance, with 62.7\% accuracy in global mode and 49.4\% in breakpoint mode, outperforming MovieChat.
Furthermore, we observe a substantial performance margin between our method and our re-implemented counterparts (+9.9\% for LLaMA-VID and +9.5\% for LLaVA-NeXT), underscoring the effectiveness of our approach towards long video scenarios.
Additionally, we conduct experiments on EgoSchema~\cite{mangalam2023egoschema} benchmark, as detailed in Table~\ref{tab-egoschema}.
Our method achieve advanced performance compared to other approaches, reaching an accuracy of 58.2\%.
The superior results on these long video benchmarks further demonstrate that our method excels in focusing on the key temporal and contextual details within video data, efficiently organizing visual tokens.
This capability enables our model to perform particularly well in long video scenarios.

To further highlight the advantages of our method in long video scenarios, we conduct additional evaluations on a diverse set of long video benchmarks, including MLVU~\cite{zhou2024mlvu}, MVBench~\cite{li2024mvbench}, LVBench~\cite{wang2024lvbench} and VideoMME~\cite{fu2024videomme}.
These benchmarks are specifically designed to assess the capability of models in understanding, reasoning, and maintaining contextual consistency across extended video sequences, making them significantly more challenging than short video benchmarks.
\noindent\textbf{MLVU:} MLVU~\cite{zhou2024mlvu} comprises a diverse collection of videos, ranging from 3 minutes to 2 hours, and include nine distinct evaluation tasks that challenge models to leverage both global and local information from videos.
\noindent\textbf{MVBench:} MVBench~\cite{li2024mvbench} is a comprehensive benchmark introduced to assess the temporal understanding capabilities of MLLMs, which covers 20 challenging video tasks.
\noindent\textbf{LVBench:} LVBench~\cite{wang2024lvbench} is specifically designed for long video understanding, which comprises publicly sourced videos, including TV series, sports broadcasts, and everyday surveillance footage, and encompasses a diverse set of tasks aimed at long video comprehension and information extraction.
\noindent\textbf{VideoMME:} VideoMME~\cite{fu2024videomme} features diversity in video types, spanning 6 primary visual domains with 30 subfields to ensure broad scenario generalizability.
The dataset includes videos of varying durations, from 11 seconds to 1 hour.

As shown in Table~\ref{tab-long4} and Table~\ref{tab-videomme}, our methods achieves outstanding performance compared with counterparts.
Notably, our advantage is particularly evident on long video-dominated datasets and evaluation metrics, \ie, LVBench and VideoMME-Long, further demonstrating the superiority of our approach in handling long video tasks.

\begin{table}[h]
\renewcommand\arraystretch{1.5}
\begin{center}
\resizebox{1.0\linewidth}{!}{
\begin{tabular}{ccc|ccccc}
\midrule[1.0pt]
Method & LLM & LoRA & MLVU & MVBench & LVBench \\
\midrule
MovieChat~\cite{song2024moviechat} & Vicuna-7B & \xmark & 25.8 & - & 22.5 \\
mPLUG-Owl-V~\cite{ye2023mplug} & LLaMA-7B & \xmark & 25.9 & 29.7 & - \\
Video-ChatGPT~\cite{maaz2023videochatgpt} & Vicuna-7B & \xmark & 31.3 & 32.7 & 23.1 \\
VideoChat~\cite{li2023videochat} & Vicuna-7B & \xmark & 29.2 & 35.5 & - \\
\midrule
LLaMA-VID & Vicuna-7B & \cmark & 46.4 & 49.7 & 21.8 \\
LLaVA-Next & Llama3-8B & \cmark & 49.8 & 51.2 & 24.2 \\
\midrule
Ours & Llama3-8B & \cmark & 55.2 & 54.5 & 29.5 \\
Ours & Qwen2-7B & \cmark & \textbf{60.1} & \textbf{60.7} & \textbf{31.3} \\
\midrule[1.0pt]
\end{tabular}}
\end{center}
\caption{Evaluation results on MLVU, MVBench and LVBench.}
\label{tab-long4}
\end{table}

\begin{table}[h]
\renewcommand\arraystretch{1.5}
\begin{center}
\resizebox{1.0\linewidth}{!}{
\begin{tabular}{ccc|cccc}
\midrule[1.0pt]
Method & LLM & LoRA & Short & Medium & Long & Overall \\
\midrule
Video-LLaVA~\cite{lin2023videollava} & Vicuna-7B & \xmark & 46.1 & 40.7 & 38.1 & 41.6 \\
VideoChat2~\cite{li2024mvbench} & Mistral-7B & \xmark & 52.8 & 39.4 & 39.2 & 43.8 \\
\midrule
LLaMA-VID & Vicuna-7B & \cmark & 46.5 & 39.3 & 39.6 & 41.8 \\
LLaVA-Next & Llama3-8B & \cmark & 52.1 & 46.3 & 44.5 & 47.6 \\
\midrule
Ours & Llama3-8B & \cmark & 60.3 & 53.6 & 51.9 & 55.3 \\
Ours & Qwen2-7B & \cmark & \textbf{61.5} & \textbf{56.5} & \textbf{54.8} & \textbf{57.6} \\
\midrule[1.0pt]
\end{tabular}}
\end{center}
\caption{Evaluation results on VideoMME.}
\label{tab-videomme}
\vspace{-4mm}
\end{table}

\subsection{Ablations}
In this section, we provide detailed ablation studies of our method conducted utilizing Llama3-8B as the foundation LLM.
More ablations are included in the \textbf{\textit{appendix}}.

\noindent\textbf{Component-wise analysis.}
We analyze the contribution of each proposed component in Table~\ref{tab-component} (Row I, II, V, VI).
To conduct a thorough exploration, ablation experiments are performed on both a short video benchmark (ActivityNet-QA) and a long video benchmark (MovieChat-1K).
Additionally, to gain insight into the visual tokens utilization across benchmarks, we report the average token number used during inference.

With the inter-frame compression strategy, we observe a substantial reduction in visual tokens.
Remarkably, performance on ActivityNet-QA even improves by +0.2\% in accuracy, while on MovieChat-1K, accuracy slightly declines by -0.5\%.
We hypothesize that MovieChat's emphasis on fine-grained movie details demands more high-level information.
Moreover, incorporating dynamic token condensation enables our method to selectively focus on frames with high temporal significance and contextual relevance, leading to an accuracy gain of 1.1\% on ActivityNet-QA and 2.5\% on MovieChat-1K (Row V).
This observation further validates our hypothesis, as demonstrated by the substantial improvements achieved with dynamic token condensation.
Finally, the integration of the spatial-temporal modeling module further enhances performance by 1.5\% and 4.8\%, respectively, highlighting the value of explicitly encoding temporal and spatial relationships of visual tokens to improve the model's comprehension ability.

\noindent\textbf{Ablation on token compression strategy.}
As shown in Table~\ref{tab-component}, to further investigate which token compression strategy performs better while eliminating the influence of token count, we fix the number of input video tokens and incrementally introduce inter-frame sampling and dynamic token compression (Row III, IV, V).
To keep token counts consistent, we randomly sample remaining frames.
Native pooling without inter-frame sampling (Row III) leads to a significant performance drop.
Introducing inter-frame sampling (Row IV) improves performance, but a gap remains compared to dynamic token compression (Row V).
Empirical results highlight the superiority of dynamic token compression over native pooling and pure inter-frame sampling.

\begin{table}[t]
\renewcommand\arraystretch{1.5}
\begin{center}
\resizebox{1.0\linewidth}{!}{
\begin{tabular}{c|cc|cc}
\midrule[1.0pt]
& \multicolumn{2}{c|}{ActivityNet-QA} & \multicolumn{2}{c}{MovieChat-1K} \\
\multirow{-2}{*}{Method} & Acc & Score & Acc & Score \\
\midrule
Katna~\cite{katna} & 45.7 & 3.1 & 49.8 & 2.4 \\
KeyVideoLLM~\cite{liang2024keyvideollm} & 46.5 & 3.1 & 50.6 & 2.5 \\
\midrule
Ours (with Katna) & 48.5 & 3.3 & 58.0 & 3.0 \\
Ours (with KeyVideoLLM) & 49.1 & 3.5 & 58.9 & 3.1 \\
\rowcolor[gray]{0.9}Ours & 49.8 & 3.6 & 60.2 & 3.2\\
\midrule[1.0pt]
\end{tabular}}
\end{center}
\caption{Comparison with alternative keyframe selection techniques by replacing our inter-frame compression component.}
\label{tab-key}
\end{table}
\begin{table}[t]
\renewcommand\arraystretch{1.5}
\begin{center}
\resizebox{1.0\linewidth}{!}{
\begin{tabular}{ccc|ccc|ccc}
\midrule[1.0pt]
\multicolumn{3}{c|}{Token condensation} & \multicolumn{3}{c|}{ActivityNet-QA} & \multicolumn{3}{c}{MovieChat-1K} \\
$\alpha$ & $d_1$ & $d_2$ & Token \# & Acc & Score & Token \# & Acc & Score \\
\midrule
0.1 & 1 & 4 & $\sim$4.0k & 49.6 & 3.5 & $\sim$7.2k & 60.2 & 3.1 \\
0.1 & 2 & 4 & $\sim$2.0k & 49.4 & 3.4 & $\sim$3.8k & 59.9 & 3.1 \\
\rowcolor[gray]{0.9}0.3 & 2 & 4 & $\sim$3.0k & 49.8 & 3.6 & $\sim$5.5k & 60.2 & 3.2 \\
0.5 & 2 & 4 & $\sim$4.0k & 50.2 & 3.5 & $\sim$7.2k & 60.3 & 3.2 \\
0.3 & 2 & 8 & $\sim$2.2k & 49.1 & 3.5 & $\sim$4.0k & 59.6 & 3.2 \\
\midrule[1.0pt]
\end{tabular}}
\end{center}
\caption{Hyper-parameters ablation. \# indicates the average visual token number on the corresponding benchmark.}
\label{tab-hyper}
\end{table}

\noindent\textbf{Comparison with alternative keyframe selection techniques.}
In addition to our approach, we notice that other keyframe selection techniques have been developed.
To provide further insights into the role of keyframe selection in video understanding, we conduct experiments with two representative methods: Katna~\cite{katna} and KeyVideoLLM~\cite{liang2024keyvideollm}.
These techniques are employed to replace the I-frame selection in our inter-frame compression component.
The results in Table~\ref{tab-key} indicate that our combined compression strategy remains effective on both benchmarks with these alternative methods integrated.
This supports the rationale of sampling video data based on its temporal property and focusing on frames of contextual importance.
Additionally, our approach significantly outperforms Katna, which is limited by its sensitivity to hyper-parameter tuning, as well as KeyVideoLLM, which, though somewhat efficient with a coarse-to-fine sampling structure, still relies on sparse sampling at its initial coarse level.
These findings underscore the efficiency and robustness of our method.

\noindent\textbf{Comparison with other token compression baselines.}
\begin{table}[h]
\renewcommand\arraystretch{1.5}
\begin{center}
\resizebox{1.0\linewidth}{!}{
\begin{tabular}{c|cccc}
\midrule[1.0pt]
Method & Anet-Acc & Anet-Score & MovieChat-Acc & MovieChat-Score \\
\midrule
naive pooling & 47.0 & 3.4 & 53.4 & 2.8 \\
\midrule
cross attn. layer & 44.8 & 3.2 & 50.5 & 2.6 \\
Q-former & 46.5 & 3.4 & 54.3 & 2.9 \\
\rowcolor[gray]{0.9}Ours & 49.8 & 3.6 & 60.2 & 3.2 \\
\midrule[1.0pt]
\end{tabular}}
\end{center}
\caption{Comparison with other token compression baselines.}
\label{tab-ablation_compression}
\vspace{-4mm}
\end{table}
We evaluate the advantages of our proposed token compression strategy by comparing it with three baseline methods: naive spatio-temporal pooling, cross-attention layer, and Q-Former.
For spatio-temporal pooling, we uniformly sample frames and downsample frame features using spatial pooling.
For the cross-attention layer and Q-Former, compressed features are generated through their respective attention mechanisms.
As shown in Table~\ref{tab-ablation_compression}, our method outperforms all baselines, demonstrating its effectiveness.

\noindent\textbf{Ablation on hybrid token condensation.}
\begin{table}[t]
\renewcommand\arraystretch{1.5}
\begin{center}
\resizebox{1.0\linewidth}{!}{
\begin{tabular}{cc|ccc|ccc}
\midrule[1.0pt]
\multicolumn{2}{c|}{Token condensation} & \multicolumn{3}{c|}{ActivityNet-QA} & \multicolumn{3}{c}{MovieChat-1K} \\
$d_1$ & $d_2$ & Token \# & Acc & Score & Token \# & Acc & Score \\
\midrule
4 & 4 & $\sim$1.9k & 47.2 & 3.3 & $\sim$3.8k & 52.9 & 2.8 \\
1 & - & $\sim$9.2k & 48.0 & 3.4 & $\sim$18.4k & - & - \\
2 & - & $\sim$2.3k & 47.8 & 3.3 & $\sim$4.6k & 51.1 & 2.5 \\
\rowcolor[gray]{0.9}2 & 4 & $\sim$3.0k & 49.8 & 3.6 & $\sim$5.5k & 60.2 & 3.2 \\
\midrule[1.0pt]
\end{tabular}}
\end{center}
\caption{Ablation on dynamic token condensation. \# indicates the average visual token number on the corresponding benchmark.}
\label{tab-dtc}
\vspace{-2mm}
\end{table}
In this section, we conduct ablations on the proposed dynamic token condensation component to investigate the effect of dynamic spatial resolutions of visual tokens.
As a comparison, we retain only the contextually relevant frames while ignoring others, applying a uniform downsample ratio $d_1$ to all frames for spatial condensation.
As shown in Table~\ref{tab-dtc} (the first row indicates the baseline), we observe that roughly removing relatively less relevant frames while maintaining a globally high resolution does not lead to better performance.
Experiments with $d_1=1$ introduce significant computational overhead, exceeding the available computational resources during inference on the MovieChat-1K dataset.
Notably, experiments with $d_1=2$ even result in worse performance than the baseline on the long-video scenarios.
We attribute this to the excessive loss of temporal information, which causes the LLM to receive primarily static content rather than dynamic details, thereby weakening its ability to achieve a comprehensive global understanding.

\noindent\textbf{Ablation on spatial-temporal modeling.}
\begin{table}[t]
\renewcommand\arraystretch{1.5}
\begin{center}
\resizebox{1.0\linewidth}{!}{
\begin{tabular}{cc|cc|cc}
\midrule[1.0pt]
\multicolumn{2}{c|}{Spatial-temporal modeling} & \multicolumn{2}{c|}{ActivityNet-QA} & \multicolumn{2}{c}{MovieChat-1K} \\
$\epsilon_{intra}$ & $\epsilon_{inter}$ & Acc & Score & Acc & Score \\
\midrule
\xmark & \xmark & 48.3 & 3.4 & 55.4 & 2.9 \\
\cmark & \xmark & 49.4 & 3.5 & 58.8 & 3.1 \\
\xmark & \cmark & 48.8 & 3.4 & 56.9 & 3.1 \\
\rowcolor[gray]{0.9}\cmark & \cmark & 49.8 & 3.6 & 60.2 & 3.2 \\
\midrule[1.0pt]
\end{tabular}}
\end{center}
\caption{Ablation on spatial-temporal modeling.}
\label{tab-st}
\vspace{-2mm}
\end{table}

We discuss the rationale behind the design of the proposed spatiotemporal modeling approach.
We first individually examine the effectiveness of discrete tokens in modeling inter-frame and intra-frame relationships.
As shown in Table~\ref{tab-st}, modeling intra-frame relationships contributes an improvement of +1.1\% on ANet-QA and +3.4\% on MovieChat-1K, while modeling inter-frame relationships provides gains of +0.5\% and +1.5\%.
Combining both enhances overall performance by +1.5\% and +4.8\%.

\begin{table}[t]
\renewcommand\arraystretch{1.5}
\begin{center}
\resizebox{1.0\linewidth}{!}{
\begin{tabular}{cc|cccc}
\midrule[1.0pt]
\multicolumn{2}{c|}{Spatial-temporal modeling} & \multicolumn{4}{c}{ActivityNet} \\
$\epsilon_{intra}$ & $\epsilon_{inter}$ & R@0.3 & R@0.5 & R@0.7 & mIoU \\
\midrule
\xmark & \xmark & 30.2 & 18.9 & 9.8 & 20.3 \\
\cmark & \xmark & 41.4 & 25.1 & 13.3 & 27.6 \\
\xmark & \cmark & 31.4 & 20.2 & 10.1 & 22.5 \\
\rowcolor[gray]{0.9}\cmark & \cmark & 42.3 & 25.8 & 13.7 & 28.2 \\
\midrule[1.0pt]
\end{tabular}}
\end{center}
\caption{Analysis of spatial-temporal modeling on temporal grounding task.}
\label{tab-st2}
\vspace{-2mm}
\end{table}

We further investigate inter-frame relationship modeling by conducting evaluations on fine-grained temporal understanding tasks.
Specifically, we perform experiments on the ActivityNet-Captions benchmark for temporal video grounding tasks.
The Intersection over Union (IoU) metric is used to measure the alignment between the time segments predicted by the model and the corresponding ground truth segments.
We report the mean IoU (mIoU) and recall for IoU thresholds (IoU $\ge m$) where $m$ is set to {0.3, 0.5, 0.7}.
As shown in Table~\ref{tab-st2}, the inclusion of inter-frame relationship modeling significantly enhances the model's performance on temporal video understanding, demonstrating its effectiveness.

\noindent\textbf{Robustness to I-frame selection configuration.}
In this section, we examine the impact of modifying I-frame compression settings.
Specifically, we adjust the Group of Pictures (GoP), a key parameter of I-frame that determines the compression ratio.
As shown in Table~\ref{tab-config}, performance remains stable across different GoP settings, demonstrating the tolerance for various configurations.

\begin{table}[t]
\renewcommand\arraystretch{1.5}
\begin{center}
\resizebox{1.0\linewidth}{!}{
\begin{tabular}{c|cccc}
\midrule[1.0pt]
GoP & Anet-Acc & Anet-Score & MovieChat-Acc & MovieChat-Score \\
\midrule
10 & 49.7 & 3.5 & 59.8 & 3.1 \\
20 & 50.1 & 3.6 & 59.8 & 3.2 \\
\rowcolor[gray]{0.9}30 & 49.8 & 3.5 & 60.2 & 3.3 \\
40 & 50.0 & 3.6 & 60.1 & 3.2 \\
\midrule[1.0pt]
\end{tabular}}
\end{center}
\caption{Robustness of I-frames compression with different GoP.}
\label{tab-config}
\vspace{-2mm}
\end{table}

\begin{table}[t]
\renewcommand\arraystretch{1.5}
\begin{center}
\resizebox{0.8\linewidth}{!}{
\begin{tabular}{c|cc|cc}
\midrule[1.0pt]
& \multicolumn{2}{c|}{ActivityNet-QA} & \multicolumn{2}{c}{MovieChat-1K} \\
\multirow{-2}{*}{LLM} & Acc & Score & Acc & Score \\
\midrule
Llama2-7B & 49.2 & 3.4 & 59.6 & 3.1 \\
Vicuna-7B & 49.4 & 3.6 & 59.5 & 3.0 \\
Llama3-8B & 49.8 & 3.6 & 60.2 & 3.2 \\
\rowcolor[gray]{0.9}Qwen2-7B & 51.2 & 3.7 & 62.7 & 3.3 \\
\midrule[1.0pt]
\end{tabular}}
\end{center}
\caption{Robustness to foundation LLM.}
\label{tab-LLM}
\vspace{-4mm}
\end{table}

\noindent\textbf{Comparison with other foundation LLMs.}
To assess the robustness of our method with respect to the foundation LLM, we replace the original base model with two alternative large language models: Llama2-7B~\cite{touvron2023llama2} and Vicuna-7B~\cite{vicuna2023}.
The results in Table~\ref{tab-LLM} indicate minimal performance variation, underscoring our method's robustness to different foundation LLMs.

\begin{figure*}[t]
    \begin{center}
        \includegraphics[width=1.0\linewidth]{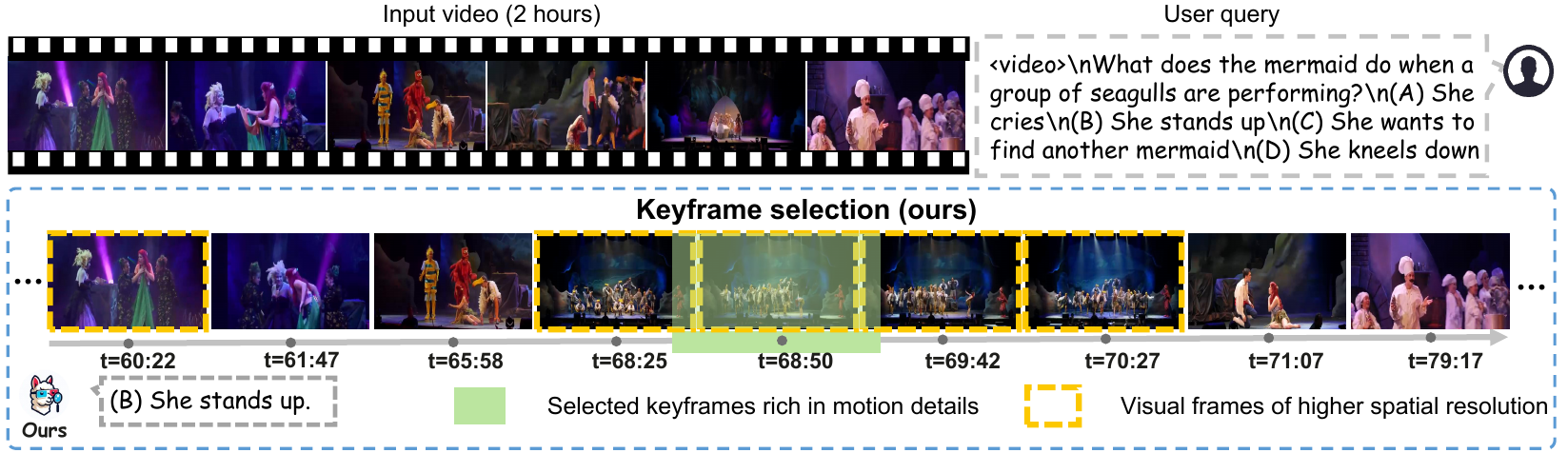}
    \end{center}
    \caption{Illustration of uniform sampling on timestamp versus keyframe selection. Uniform sampling on long video sequences risks both frame redundancy and missing motion details, while our keyframe selection reduces redundancy and more effectively captures motion dynamics.}
    \label{fig-appvis1}
\end{figure*}

\begin{figure*}[t]
    \begin{center}
        \includegraphics[width=1.0\linewidth]{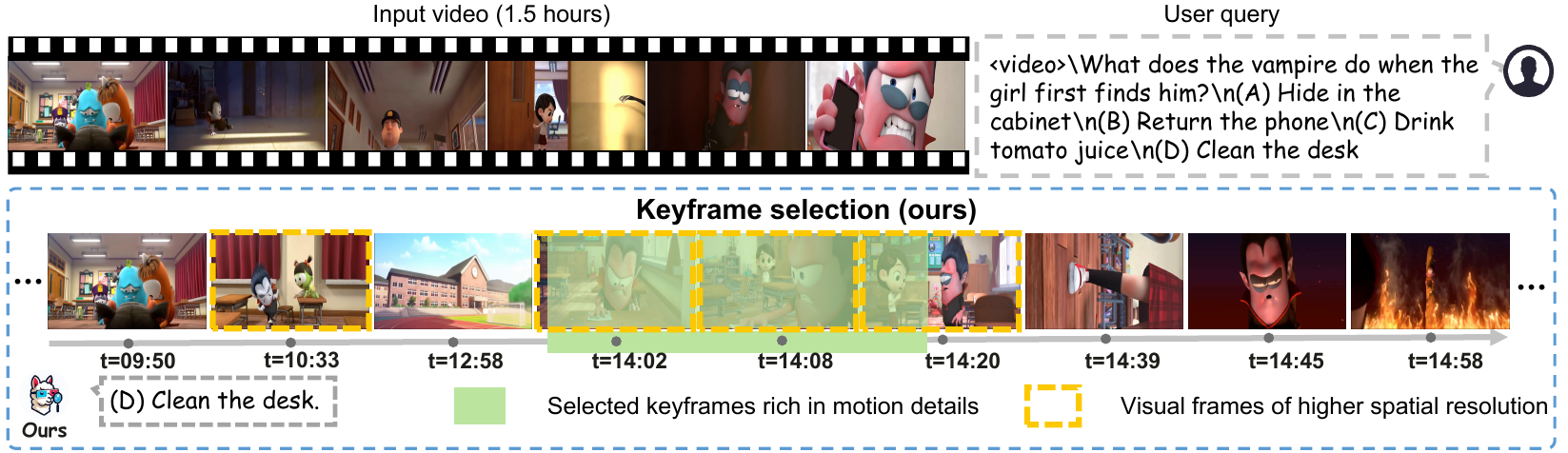}
    \end{center}
    \caption{Illustration of uniform sampling on timestamp versus keyframe selection. Uniform sampling on long video sequences risks both frame redundancy and missing motion details, while our keyframe selection reduces redundancy and more effectively captures motion dynamics.}
    \label{fig-appvis2}
\end{figure*}

\noindent\textbf{Hyper-parameters ablation.}
We conduct ablation studies to identify optimal hyper-parameter settings in our proposed token condensation component.
As shown in Table~\ref{tab-hyper}, adjusting the contextual focusing ratio $\alpha$ from 0.1 to 0.5 results in minimal performance variation, suggesting that temporal redundancy is prevalent even within semantically informative frames, thus supporting the robustness of our method to changes in $\alpha$.
Next, we vary the token condensation ratio $d_1$ from 2 to 1.
This change yields a slight performance increase but a significant rise in visual token count (nearly double the visual tokens), indicating that maintaining a high spatial resolution of 256 tokens offers limited advantages.
To balance efficiency and performance, we set $d_1$ to 2.
As increasing $d_2$ from 4 to 8 slightly reduces performance, we set $d_2$ at 4 to maintain comprehension quality.

\subsection{Visualizations}

Figure~\ref{fig-appvis1} and Figure~\ref{fig-appvis2} illustrate the qualitative results of our method on long video scenarios.
Specifically, we observe that:
(i) The non-uniform keyframe sampling accurately captures complex motion changes, as indicated by the green markers.
(ii) Dynamic token condensation effectively identifies query-related visual features and allocates higher spatial resolution to them, enhancing their representational capacity, as shown by the yellow markers.
These findings further validate the effectiveness of our approach.
\section{Conclusion}
We introduce KFFocus, a multimodal large language model tailored for fine-grained spatiotemporal understanding in video modality.
We revisit the inherent limitations of uniform sampling on video data in existing methods, which restrict the model's ability to efficiently capture informative details essential for accurate comprehension.
To tackle this issue, we propose a unified inter-frame and intra-frame compression framework to address visual token redundancy while preserving crucial temporal and semantic information.
Our proposed hybrid token compression enhances the model's ability to selectively retain informative tokens across frames, optimizing both efficiency and accuracy.
Additionally, our newly developed spatial-temporal modeling module explicitly encodes temporal dynamics and spatial relationships, further strengthening the capacity for nuanced video comprehension.
Demonstrating superior performance on various video understanding benchmarks, we hope that our work can propel the development of Vid-LLMs towards advanced video understanding capabilities.

\section{Data Availability Statement}
The datasets generated during and/or analysed during the current study are available in the MSVD-QA~\cite{xu2017video}, MSRVTT-QA~\cite{xu2016msr}, ActivityNet-QA~\cite{yu2019activitynet}, video-based generative benchmark~\cite{maaz2023videochatgpt}, MovieChat-1K~\cite{song2024moviechat}, EgoSchema~\cite{mangalam2023egoschema}, MLVU~\cite{zhou2024mlvu}, MVBench~\cite{li2024mvbench}, LVBench~\cite{wang2024lvbench} and VideoMME~\cite{fu2024videomme}.

\bibliography{sn-bibliography}


\begin{thebibliography}{47}
\ifx \bisbn   \undefined \def \bisbn  #1{ISBN #1}\fi
\ifx \binits  \undefined \def \binits#1{#1}\fi
\ifx \bauthor  \undefined \def \bauthor#1{#1}\fi
\ifx \batitle  \undefined \def \batitle#1{#1}\fi
\ifx \bjtitle  \undefined \def \bjtitle#1{#1}\fi
\ifx \bvolume  \undefined \def \bvolume#1{\textbf{#1}}\fi
\ifx \byear  \undefined \def \byear#1{#1}\fi
\ifx \bissue  \undefined \def \bissue#1{#1}\fi
\ifx \bfpage  \undefined \def \bfpage#1{#1}\fi
\ifx \blpage  \undefined \def \blpage #1{#1}\fi
\ifx \burl  \undefined \def \burl#1{\textsf{#1}}\fi
\ifx \doiurl  \undefined \def \doiurl#1{\url{https://doi.org/#1}}\fi
\ifx \betal  \undefined \def \betal{\textit{et al.}}\fi
\ifx \binstitute  \undefined \def \binstitute#1{#1}\fi
\ifx \binstitutionaled  \undefined \def \binstitutionaled#1{#1}\fi
\ifx \bctitle  \undefined \def \bctitle#1{#1}\fi
\ifx \beditor  \undefined \def \beditor#1{#1}\fi
\ifx \bpublisher  \undefined \def \bpublisher#1{#1}\fi
\ifx \bbtitle  \undefined \def \bbtitle#1{#1}\fi
\ifx \bedition  \undefined \def \bedition#1{#1}\fi
\ifx \bseriesno  \undefined \def \bseriesno#1{#1}\fi
\ifx \blocation  \undefined \def \blocation#1{#1}\fi
\ifx \bsertitle  \undefined \def \bsertitle#1{#1}\fi
\ifx \bsnm \undefined \def \bsnm#1{#1}\fi
\ifx \bsuffix \undefined \def \bsuffix#1{#1}\fi
\ifx \bparticle \undefined \def \bparticle#1{#1}\fi
\ifx \barticle \undefined \def \barticle#1{#1}\fi
\bibcommenthead
\ifx \bconfdate \undefined \def \bconfdate #1{#1}\fi
\ifx \botherref \undefined \def \botherref #1{#1}\fi
\ifx \url \undefined \def \url#1{\textsf{#1}}\fi
\ifx \bchapter \undefined \def \bchapter#1{#1}\fi
\ifx \bbook \undefined \def \bbook#1{#1}\fi
\ifx \bcomment \undefined \def \bcomment#1{#1}\fi
\ifx \oauthor \undefined \def \oauthor#1{#1}\fi
\ifx \citeauthoryear \undefined \def \citeauthoryear#1{#1}\fi
\ifx \endbibitem  \undefined \def \endbibitem {}\fi
\ifx \bconflocation  \undefined \def \bconflocation#1{#1}\fi
\ifx \arxivurl  \undefined \def \arxivurl#1{\textsf{#1}}\fi
\csname PreBibitemsHook\endcsname

\bibitem[\protect\citeauthoryear{Li et~al.}{2023}]{li2023videochat}
\begin{botherref}
\oauthor{\bsnm{Li}, \binits{K.}},
\oauthor{\bsnm{He}, \binits{Y.}},
\oauthor{\bsnm{Wang}, \binits{Y.}},
\oauthor{\bsnm{Li}, \binits{Y.}},
\oauthor{\bsnm{Wang}, \binits{W.}},
\oauthor{\bsnm{Luo}, \binits{P.}},
\oauthor{\bsnm{Wang}, \binits{Y.}},
\oauthor{\bsnm{Wang}, \binits{L.}},
\oauthor{\bsnm{Qiao}, \binits{Y.}}:
Videochat: Chat-centric video understanding.
arXiv preprint
(2023)
\end{botherref}
\endbibitem

\bibitem[\protect\citeauthoryear{Lin et~al.}{2023}]{lin2023videollava}
\begin{botherref}
\oauthor{\bsnm{Lin}, \binits{B.}},
\oauthor{\bsnm{Zhu}, \binits{B.}},
\oauthor{\bsnm{Ye}, \binits{Y.}},
\oauthor{\bsnm{Ning}, \binits{M.}},
\oauthor{\bsnm{Jin}, \binits{P.}},
\oauthor{\bsnm{Yuan}, \binits{L.}}:
Video-llava: Learning united visual representation by alignment before projection.
arXiv preprint
(2023)
\end{botherref}
\endbibitem

\bibitem[\protect\citeauthoryear{Zhang et~al.}{2023}]{zhang2023videollama}
\begin{botherref}
\oauthor{\bsnm{Zhang}, \binits{H.}},
\oauthor{\bsnm{Li}, \binits{X.}},
\oauthor{\bsnm{Bing}, \binits{L.}}:
Video-llama: An instruction-tuned audio-visual language model for video understanding.
arXiv preprint
(2023)
\end{botherref}
\endbibitem

\bibitem[\protect\citeauthoryear{Li et~al.}{2023}]{li2023blip}
\begin{bchapter}
\bauthor{\bsnm{Li}, \binits{J.}},
\bauthor{\bsnm{Li}, \binits{D.}},
\bauthor{\bsnm{Savarese}, \binits{S.}},
\bauthor{\bsnm{Hoi}, \binits{S.}}:
\bctitle{Blip-2: Bootstrapping language-image pre-training with frozen image encoders and large language models}.
In: \bbtitle{ICML}
(\byear{2023})
\end{bchapter}
\endbibitem

\bibitem[\protect\citeauthoryear{Zhu et~al.}{2023}]{zhu2023minigpt}
\begin{botherref}
\oauthor{\bsnm{Zhu}, \binits{D.}},
\oauthor{\bsnm{Chen}, \binits{J.}},
\oauthor{\bsnm{Shen}, \binits{X.}},
\oauthor{\bsnm{Li}, \binits{X.}},
\oauthor{\bsnm{Elhoseiny}, \binits{M.}}:
Minigpt-4: Enhancing vision-language understanding with advanced large language models.
arXiv preprint
(2023)
\end{botherref}
\endbibitem

\bibitem[\protect\citeauthoryear{Liu et~al.}{2024a}]{liu2024visual}
\begin{botherref}
\oauthor{\bsnm{Liu}, \binits{H.}},
\oauthor{\bsnm{Li}, \binits{C.}},
\oauthor{\bsnm{Wu}, \binits{Q.}},
\oauthor{\bsnm{Lee}, \binits{Y.J.}}:
Visual instruction tuning.
Advances in neural information processing systems
(2024)
\end{botherref}
\endbibitem

\bibitem[\protect\citeauthoryear{Liu et~al.}{2024b}]{liu2024llavanext}
\begin{botherref}
\oauthor{\bsnm{Liu}, \binits{H.}},
\oauthor{\bsnm{Li}, \binits{C.}},
\oauthor{\bsnm{Li}, \binits{Y.}},
\oauthor{\bsnm{Li}, \binits{B.}},
\oauthor{\bsnm{Zhang}, \binits{Y.}},
\oauthor{\bsnm{Shen}, \binits{S.}},
\oauthor{\bsnm{Lee}, \binits{Y.J.}}:
LLaVA-NeXT: Improved reasoning, OCR, and world knowledge
(2024).
\url{https://llava-vl.github.io/blog/2024-01-30-llava-next/}
\end{botherref}
\endbibitem

\bibitem[\protect\citeauthoryear{Peng et~al.}{2023}]{peng2023kosmos}
\begin{botherref}
\oauthor{\bsnm{Peng}, \binits{Z.}},
\oauthor{\bsnm{Wang}, \binits{W.}},
\oauthor{\bsnm{Dong}, \binits{L.}},
\oauthor{\bsnm{Hao}, \binits{Y.}},
\oauthor{\bsnm{Huang}, \binits{S.}},
\oauthor{\bsnm{Ma}, \binits{S.}},
\oauthor{\bsnm{Wei}, \binits{F.}}:
Kosmos-2: Grounding multimodal large language models to the world.
arXiv preprint
(2023)
\end{botherref}
\endbibitem

\bibitem[\protect\citeauthoryear{Wang et~al.}{2024}]{wang2024visionllm}
\begin{botherref}
\oauthor{\bsnm{Wang}, \binits{W.}},
\oauthor{\bsnm{Chen}, \binits{Z.}},
\oauthor{\bsnm{Chen}, \binits{X.}},
\oauthor{\bsnm{Wu}, \binits{J.}},
\oauthor{\bsnm{Zhu}, \binits{X.}},
\oauthor{\bsnm{Zeng}, \binits{G.}},
\oauthor{\bsnm{Luo}, \binits{P.}},
\oauthor{\bsnm{Lu}, \binits{T.}},
\oauthor{\bsnm{Zhou}, \binits{J.}},
\oauthor{\bsnm{Qiao}, \binits{Y.}}, et al.:
Visionllm: Large language model is also an open-ended decoder for vision-centric tasks.
Advances in Neural Information Processing Systems
(2024)
\end{botherref}
\endbibitem

\bibitem[\protect\citeauthoryear{Maaz et~al.}{2023}]{maaz2023videochatgpt}
\begin{botherref}
\oauthor{\bsnm{Maaz}, \binits{M.}},
\oauthor{\bsnm{Rasheed}, \binits{H.}},
\oauthor{\bsnm{Khan}, \binits{S.}},
\oauthor{\bsnm{Khan}, \binits{F.S.}}:
Video-chatgpt: Towards detailed video understanding via large vision and language models.
arXiv preprint
(2023)
\end{botherref}
\endbibitem

\bibitem[\protect\citeauthoryear{Wang et~al.}{2023}]{wang2023internvid}
\begin{botherref}
\oauthor{\bsnm{Wang}, \binits{Y.}},
\oauthor{\bsnm{He}, \binits{Y.}},
\oauthor{\bsnm{Li}, \binits{Y.}},
\oauthor{\bsnm{Li}, \binits{K.}},
\oauthor{\bsnm{Yu}, \binits{J.}},
\oauthor{\bsnm{Ma}, \binits{X.}},
\oauthor{\bsnm{Li}, \binits{X.}},
\oauthor{\bsnm{Chen}, \binits{G.}},
\oauthor{\bsnm{Chen}, \binits{X.}},
\oauthor{\bsnm{Wang}, \binits{Y.}}, et al.:
Internvid: A large-scale video-text dataset for multimodal understanding and generation.
arXiv preprint
(2023)
\end{botherref}
\endbibitem

\bibitem[\protect\citeauthoryear{Wang et~al.}{2024}]{wang2024internvideo2}
\begin{botherref}
\oauthor{\bsnm{Wang}, \binits{Y.}},
\oauthor{\bsnm{Li}, \binits{K.}},
\oauthor{\bsnm{Li}, \binits{X.}},
\oauthor{\bsnm{Yu}, \binits{J.}},
\oauthor{\bsnm{He}, \binits{Y.}},
\oauthor{\bsnm{Chen}, \binits{G.}},
\oauthor{\bsnm{Pei}, \binits{B.}},
\oauthor{\bsnm{Zheng}, \binits{R.}},
\oauthor{\bsnm{Xu}, \binits{J.}},
\oauthor{\bsnm{Wang}, \binits{Z.}}, et al.:
Internvideo2: Scaling video foundation models for multimodal video understanding.
arXiv preprint
(2024)
\end{botherref}
\endbibitem

\bibitem[\protect\citeauthoryear{Ataallah et~al.}{2024}]{ataallah2024minigpt4video}
\begin{botherref}
\oauthor{\bsnm{Ataallah}, \binits{K.}},
\oauthor{\bsnm{Shen}, \binits{X.}},
\oauthor{\bsnm{Abdelrahman}, \binits{E.}},
\oauthor{\bsnm{Sleiman}, \binits{E.}},
\oauthor{\bsnm{Zhu}, \binits{D.}},
\oauthor{\bsnm{Ding}, \binits{J.}},
\oauthor{\bsnm{Elhoseiny}, \binits{M.}}:
Minigpt4-video: Advancing multimodal llms for video understanding with interleaved visual-textual tokens.
arXiv preprint
(2024)
\end{botherref}
\endbibitem

\bibitem[\protect\citeauthoryear{KeplerLab}{2019}]{katna}
\begin{botherref}
\oauthor{\bsnm{KeplerLab}}:
Katna: Tool for automating video keyframe extraction, video compression, image autocrop and smart image resize tasks.
https://github.com/keplerlab/katna
(2019)
\end{botherref}
\endbibitem

\bibitem[\protect\citeauthoryear{Jin et~al.}{2024}]{jin2024video}
\begin{botherref}
\oauthor{\bsnm{Jin}, \binits{Y.}},
\oauthor{\bsnm{Sun}, \binits{Z.}},
\oauthor{\bsnm{Xu}, \binits{K.}},
\oauthor{\bsnm{Chen}, \binits{L.}},
\oauthor{\bsnm{Jiang}, \binits{H.}},
\oauthor{\bsnm{Huang}, \binits{Q.}},
\oauthor{\bsnm{Song}, \binits{C.}},
\oauthor{\bsnm{Liu}, \binits{Y.}},
\oauthor{\bsnm{Zhang}, \binits{D.}},
\oauthor{\bsnm{Song}, \binits{Y.}}, et al.:
Video-lavit: Unified video-language pre-training with decoupled visual-motional tokenization.
arXiv preprint
(2024)
\end{botherref}
\endbibitem

\bibitem[\protect\citeauthoryear{Liang et~al.}{2024}]{liang2024keyvideollm}
\begin{botherref}
\oauthor{\bsnm{Liang}, \binits{H.}},
\oauthor{\bsnm{Li}, \binits{J.}},
\oauthor{\bsnm{Bai}, \binits{T.}},
\oauthor{\bsnm{Chen}, \binits{C.}},
\oauthor{\bsnm{He}, \binits{C.}},
\oauthor{\bsnm{Cui}, \binits{B.}},
\oauthor{\bsnm{Zhang}, \binits{W.}}:
Keyvideollm: Towards large-scale video keyframe selection.
arXiv preprint
(2024)
\end{botherref}
\endbibitem

\bibitem[\protect\citeauthoryear{Sikora}{1997}]{sikora1997mpeg}
\begin{botherref}
\oauthor{\bsnm{Sikora}, \binits{T.}}:
The mpeg-4 video standard verification model.
IEEE Transactions on circuits and systems for video technology
(1997)
\end{botherref}
\endbibitem

\bibitem[\protect\citeauthoryear{Radford et~al.}{2021}]{radford2021learning}
\begin{bchapter}
\bauthor{\bsnm{Radford}, \binits{A.}},
\bauthor{\bsnm{Kim}, \binits{J.W.}},
\bauthor{\bsnm{Hallacy}, \binits{C.}},
\bauthor{\bsnm{Ramesh}, \binits{A.}},
\bauthor{\bsnm{Goh}, \binits{G.}},
\bauthor{\bsnm{Agarwal}, \binits{S.}},
\bauthor{\bsnm{Sastry}, \binits{G.}},
\bauthor{\bsnm{Askell}, \binits{A.}},
\bauthor{\bsnm{Mishkin}, \binits{P.}},
\bauthor{\bsnm{Clark}, \binits{J.}}, \betal:
\bctitle{Learning transferable visual models from natural language supervision}.
In: \bbtitle{International Conference on Machine Learning}
(\byear{2021})
\end{bchapter}
\endbibitem

\bibitem[\protect\citeauthoryear{Xu et~al.}{2024}]{xu2024llavauhd}
\begin{botherref}
\oauthor{\bsnm{Xu}, \binits{R.}},
\oauthor{\bsnm{Yao}, \binits{Y.}},
\oauthor{\bsnm{Guo}, \binits{Z.}},
\oauthor{\bsnm{Cui}, \binits{J.}},
\oauthor{\bsnm{Ni}, \binits{Z.}},
\oauthor{\bsnm{Ge}, \binits{C.}},
\oauthor{\bsnm{Chua}, \binits{T.-S.}},
\oauthor{\bsnm{Liu}, \binits{Z.}},
\oauthor{\bsnm{Sun}, \binits{M.}},
\oauthor{\bsnm{Huang}, \binits{G.}}:
Llava-uhd: an lmm perceiving any aspect ratio and high-resolution images.
arXiv preprint
(2024)
\end{botherref}
\endbibitem

\bibitem[\protect\citeauthoryear{Li et~al.}{2023}]{li2023llamavid}
\begin{botherref}
\oauthor{\bsnm{Li}, \binits{Y.}},
\oauthor{\bsnm{Wang}, \binits{C.}},
\oauthor{\bsnm{Jia}, \binits{J.}}:
Llama-vid: An image is worth 2 tokens in large language models.
arXiv preprint
(2023)
\end{botherref}
\endbibitem

\bibitem[\protect\citeauthoryear{Bain et~al.}{2021}]{bain2021frozen}
\begin{bchapter}
\bauthor{\bsnm{Bain}, \binits{M.}},
\bauthor{\bsnm{Nagrani}, \binits{A.}},
\bauthor{\bsnm{Varol}, \binits{G.}},
\bauthor{\bsnm{Zisserman}, \binits{A.}}:
\bctitle{Frozen in time: A joint video and image encoder for end-to-end retrieval}.
In: \bbtitle{ICCV}
(\byear{2021})
\end{bchapter}
\endbibitem

\bibitem[\protect\citeauthoryear{Yang et~al.}{2022}]{yang2022zero}
\begin{botherref}
\oauthor{\bsnm{Yang}, \binits{A.}},
\oauthor{\bsnm{Miech}, \binits{A.}},
\oauthor{\bsnm{Sivic}, \binits{J.}},
\oauthor{\bsnm{Laptev}, \binits{I.}},
\oauthor{\bsnm{Schmid}, \binits{C.}}:
Zero-shot video question answering via frozen bidirectional language models.
Advances in Neural Information Processing Systems
(2022)
\end{botherref}
\endbibitem

\bibitem[\protect\citeauthoryear{Zhang et~al.}{2023}]{zhang2023llama}
\begin{botherref}
\oauthor{\bsnm{Zhang}, \binits{R.}},
\oauthor{\bsnm{Han}, \binits{J.}},
\oauthor{\bsnm{Liu}, \binits{C.}},
\oauthor{\bsnm{Gao}, \binits{P.}},
\oauthor{\bsnm{Zhou}, \binits{A.}},
\oauthor{\bsnm{Hu}, \binits{X.}},
\oauthor{\bsnm{Yan}, \binits{S.}},
\oauthor{\bsnm{Lu}, \binits{P.}},
\oauthor{\bsnm{Li}, \binits{H.}},
\oauthor{\bsnm{Qiao}, \binits{Y.}}:
Llama-adapter: Efficient fine-tuning of language models with zero-init attention.
arXiv preprint
(2023)
\end{botherref}
\endbibitem

\bibitem[\protect\citeauthoryear{Huang et~al.}{2023}]{huang2023vtimellm}
\begin{botherref}
\oauthor{\bsnm{Huang}, \binits{B.}},
\oauthor{\bsnm{Wang}, \binits{X.}},
\oauthor{\bsnm{Chen}, \binits{H.}},
\oauthor{\bsnm{Song}, \binits{Z.}},
\oauthor{\bsnm{Zhu}, \binits{W.}}:
Vtimellm: Empower llm to grasp video moments.
arXiv preprint
(2023)
\end{botherref}
\endbibitem

\bibitem[\protect\citeauthoryear{Hu et~al.}{2021}]{hu2021lora}
\begin{botherref}
\oauthor{\bsnm{Hu}, \binits{E.J.}},
\oauthor{\bsnm{Shen}, \binits{Y.}},
\oauthor{\bsnm{Wallis}, \binits{P.}},
\oauthor{\bsnm{Allen-Zhu}, \binits{Z.}},
\oauthor{\bsnm{Li}, \binits{Y.}},
\oauthor{\bsnm{Wang}, \binits{S.}},
\oauthor{\bsnm{Wang}, \binits{L.}},
\oauthor{\bsnm{Chen}, \binits{W.}}:
Lora: Low-rank adaptation of large language models.
arXiv preprint
(2021)
\end{botherref}
\endbibitem

\bibitem[\protect\citeauthoryear{Hore and Ziou}{2010}]{hore2010image-psnr}
\begin{bchapter}
\bauthor{\bsnm{Hore}, \binits{A.}},
\bauthor{\bsnm{Ziou}, \binits{D.}}:
\bctitle{Image quality metrics: Psnr vs. ssim}.
In: \bbtitle{2010 20th International Conference on Pattern Recognition}
(\byear{2010})
\end{bchapter}
\endbibitem

\bibitem[\protect\citeauthoryear{Song et~al.}{2024}]{song2024moviechat}
\begin{bchapter}
\bauthor{\bsnm{Song}, \binits{E.}},
\bauthor{\bsnm{Chai}, \binits{W.}},
\bauthor{\bsnm{Wang}, \binits{G.}},
\bauthor{\bsnm{Zhang}, \binits{Y.}},
\bauthor{\bsnm{Zhou}, \binits{H.}},
\bauthor{\bsnm{Wu}, \binits{F.}},
\bauthor{\bsnm{Chi}, \binits{H.}},
\bauthor{\bsnm{Guo}, \binits{X.}},
\bauthor{\bsnm{Ye}, \binits{T.}},
\bauthor{\bsnm{Zhang}, \binits{Y.}}, \betal:
\bctitle{Moviechat: From dense token to sparse memory for long video understanding}.
In: \bbtitle{CVPR}
(\byear{2024})
\end{bchapter}
\endbibitem

\bibitem[\protect\citeauthoryear{Yang et~al.}{2022}]{frozenbilm}
\begin{botherref}
\oauthor{\bsnm{Yang}, \binits{A.}},
\oauthor{\bsnm{Miech}, \binits{A.}},
\oauthor{\bsnm{Sivic}, \binits{J.}},
\oauthor{\bsnm{Laptev}, \binits{I.}},
\oauthor{\bsnm{Schmid}, \binits{C.}}:
Zero-shot video question answering via frozen bidirectional language models.
Advances in Neural Information Processing Systems
(2022)
\end{botherref}
\endbibitem

\bibitem[\protect\citeauthoryear{He et~al.}{2020}]{he2020deberta}
\begin{botherref}
\oauthor{\bsnm{He}, \binits{P.}},
\oauthor{\bsnm{Liu}, \binits{X.}},
\oauthor{\bsnm{Gao}, \binits{J.}},
\oauthor{\bsnm{Chen}, \binits{W.}}:
Deberta: Decoding-enhanced bert with disentangled attention.
arXiv preprint arXiv:2006.03654
(2020)
\end{botherref}
\endbibitem

\bibitem[\protect\citeauthoryear{Fu et~al.}{2021}]{fu2021violet}
\begin{botherref}
\oauthor{\bsnm{Fu}, \binits{T.-J.}},
\oauthor{\bsnm{Li}, \binits{L.}},
\oauthor{\bsnm{Gan}, \binits{Z.}},
\oauthor{\bsnm{Lin}, \binits{K.}},
\oauthor{\bsnm{Wang}, \binits{W.Y.}},
\oauthor{\bsnm{Wang}, \binits{L.}},
\oauthor{\bsnm{Liu}, \binits{Z.}}:
Violet: End-to-end video-language transformers with masked visual-token modeling.
arXiv preprint
(2021)
\end{botherref}
\endbibitem

\bibitem[\protect\citeauthoryear{Wang et~al.}{2022}]{wang2022internvideo}
\begin{botherref}
\oauthor{\bsnm{Wang}, \binits{Y.}},
\oauthor{\bsnm{Li}, \binits{K.}},
\oauthor{\bsnm{Li}, \binits{Y.}},
\oauthor{\bsnm{He}, \binits{Y.}},
\oauthor{\bsnm{Huang}, \binits{B.}},
\oauthor{\bsnm{Zhao}, \binits{Z.}},
\oauthor{\bsnm{Zhang}, \binits{H.}},
\oauthor{\bsnm{Xu}, \binits{J.}},
\oauthor{\bsnm{Liu}, \binits{Y.}},
\oauthor{\bsnm{Wang}, \binits{Z.}}, et al.:
Internvideo: General video foundation models via generative and discriminative learning.
arXiv preprint
(2022)
\end{botherref}
\endbibitem

\bibitem[\protect\citeauthoryear{Zhang et~al.}{2023}]{zhang2023llovi}
\begin{botherref}
\oauthor{\bsnm{Zhang}, \binits{C.}},
\oauthor{\bsnm{Lu}, \binits{T.}},
\oauthor{\bsnm{Islam}, \binits{M.M.}},
\oauthor{\bsnm{Wang}, \binits{Z.}},
\oauthor{\bsnm{Yu}, \binits{S.}},
\oauthor{\bsnm{Bansal}, \binits{M.}},
\oauthor{\bsnm{Bertasius}, \binits{G.}}:
A simple llm framework for long-range video question-answering.
arXiv preprint
(2023)
\end{botherref}
\endbibitem

\bibitem[\protect\citeauthoryear{Touvron et~al.}{2023}]{touvron2023llama2}
\begin{botherref}
\oauthor{\bsnm{Touvron}, \binits{H.}},
\oauthor{\bsnm{Martin}, \binits{L.}},
\oauthor{\bsnm{Stone}, \binits{K.}},
\oauthor{\bsnm{Albert}, \binits{P.}},
\oauthor{\bsnm{Almahairi}, \binits{A.}},
\oauthor{\bsnm{Babaei}, \binits{Y.}},
\oauthor{\bsnm{Bashlykov}, \binits{N.}},
\oauthor{\bsnm{Batra}, \binits{S.}},
\oauthor{\bsnm{Bhargava}, \binits{P.}},
\oauthor{\bsnm{Bhosale}, \binits{S.}}, et al.:
Llama 2: Open foundation and fine-tuned chat models.
arXiv preprint
(2023)
\end{botherref}
\endbibitem

\bibitem[\protect\citeauthoryear{Wang et~al.}{2023}]{wang2023vamos}
\begin{botherref}
\oauthor{\bsnm{Wang}, \binits{S.}},
\oauthor{\bsnm{Zhao}, \binits{Q.}},
\oauthor{\bsnm{Do}, \binits{M.Q.}},
\oauthor{\bsnm{Agarwal}, \binits{N.}},
\oauthor{\bsnm{Lee}, \binits{K.}},
\oauthor{\bsnm{Sun}, \binits{C.}}:
Vamos: Versatile action models for video understanding.
arXiv preprint
(2023)
\end{botherref}
\endbibitem

\bibitem[\protect\citeauthoryear{Touvron et~al.}{2023}]{touvron2023llama}
\begin{botherref}
\oauthor{\bsnm{Touvron}, \binits{H.}},
\oauthor{\bsnm{Lavril}, \binits{T.}},
\oauthor{\bsnm{Izacard}, \binits{G.}},
\oauthor{\bsnm{Martinet}, \binits{X.}},
\oauthor{\bsnm{Lachaux}, \binits{M.-A.}},
\oauthor{\bsnm{Lacroix}, \binits{T.}},
\oauthor{\bsnm{Rozi{\`e}re}, \binits{B.}},
\oauthor{\bsnm{Goyal}, \binits{N.}},
\oauthor{\bsnm{Hambro}, \binits{E.}},
\oauthor{\bsnm{Azhar}, \binits{F.}}, et al.:
Llama: Open and efficient foundation language models.
arXiv preprint
(2023)
\end{botherref}
\endbibitem

\bibitem[\protect\citeauthoryear{Yang et~al.}{2024}]{yang2024qwen2}
\begin{botherref}
\oauthor{\bsnm{Yang}, \binits{A.}},
\oauthor{\bsnm{Yang}, \binits{B.}},
\oauthor{\bsnm{Zhang}, \binits{B.}},
\oauthor{\bsnm{Hui}, \binits{B.}},
\oauthor{\bsnm{Zheng}, \binits{B.}},
\oauthor{\bsnm{Yu}, \binits{B.}},
\oauthor{\bsnm{Li}, \binits{C.}},
\oauthor{\bsnm{Liu}, \binits{D.}},
\oauthor{\bsnm{Huang}, \binits{F.}},
\oauthor{\bsnm{Wei}, \binits{H.}}, et al.:
Qwen2. 5 technical report.
arXiv preprint
(2024)
\end{botherref}
\endbibitem

\bibitem[\protect\citeauthoryear{Loshchilov and Hutter}{2017}]{loshchilov2017decoupled}
\begin{botherref}
\oauthor{\bsnm{Loshchilov}, \binits{I.}},
\oauthor{\bsnm{Hutter}, \binits{F.}}:
Decoupled weight decay regularization.
arXiv preprint
(2017)
\end{botherref}
\endbibitem

\bibitem[\protect\citeauthoryear{Xu et~al.}{2017}]{xu2017video}
\begin{bchapter}
\bauthor{\bsnm{Xu}, \binits{D.}},
\bauthor{\bsnm{Zhao}, \binits{Z.}},
\bauthor{\bsnm{Xiao}, \binits{J.}},
\bauthor{\bsnm{Wu}, \binits{F.}},
\bauthor{\bsnm{Zhang}, \binits{H.}},
\bauthor{\bsnm{He}, \binits{X.}},
\bauthor{\bsnm{Zhuang}, \binits{Y.}}:
\bctitle{Video question answering via gradually refined attention over appearance and motion}.
In: \bbtitle{ACM MM}
(\byear{2017})
\end{bchapter}
\endbibitem

\bibitem[\protect\citeauthoryear{Xu et~al.}{2016}]{xu2016msr}
\begin{bchapter}
\bauthor{\bsnm{Xu}, \binits{J.}},
\bauthor{\bsnm{Mei}, \binits{T.}},
\bauthor{\bsnm{Yao}, \binits{T.}},
\bauthor{\bsnm{Rui}, \binits{Y.}}:
\bctitle{Msr-vtt: A large video description dataset for bridging video and language}.
In: \bbtitle{CVPR}
(\byear{2016})
\end{bchapter}
\endbibitem

\bibitem[\protect\citeauthoryear{Yu et~al.}{2019}]{yu2019activitynet}
\begin{bchapter}
\bauthor{\bsnm{Yu}, \binits{Z.}},
\bauthor{\bsnm{Xu}, \binits{D.}},
\bauthor{\bsnm{Yu}, \binits{J.}},
\bauthor{\bsnm{Yu}, \binits{T.}},
\bauthor{\bsnm{Zhao}, \binits{Z.}},
\bauthor{\bsnm{Zhuang}, \binits{Y.}},
\bauthor{\bsnm{Tao}, \binits{D.}}:
\bctitle{Activitynet-qa: A dataset for understanding complex web videos via question answering}.
In: \bbtitle{AAAI}
(\byear{2019})
\end{bchapter}
\endbibitem

\bibitem[\protect\citeauthoryear{Mangalam et~al.}{2023}]{mangalam2023egoschema}
\begin{botherref}
\oauthor{\bsnm{Mangalam}, \binits{K.}},
\oauthor{\bsnm{Akshulakov}, \binits{R.}},
\oauthor{\bsnm{Malik}, \binits{J.}}:
Egoschema: A diagnostic benchmark for very long-form video language understanding.
Advances in Neural Information Processing Systems
(2023)
\end{botherref}
\endbibitem

\bibitem[\protect\citeauthoryear{Zhou et~al.}{2024}]{zhou2024mlvu}
\begin{botherref}
\oauthor{\bsnm{Zhou}, \binits{J.}},
\oauthor{\bsnm{Shu}, \binits{Y.}},
\oauthor{\bsnm{Zhao}, \binits{B.}},
\oauthor{\bsnm{Wu}, \binits{B.}},
\oauthor{\bsnm{Xiao}, \binits{S.}},
\oauthor{\bsnm{Yang}, \binits{X.}},
\oauthor{\bsnm{Xiong}, \binits{Y.}},
\oauthor{\bsnm{Zhang}, \binits{B.}},
\oauthor{\bsnm{Huang}, \binits{T.}},
\oauthor{\bsnm{Liu}, \binits{Z.}}:
Mlvu: A comprehensive benchmark for multi-task long video understanding.
arXiv preprint
(2024)
\end{botherref}
\endbibitem

\bibitem[\protect\citeauthoryear{Li et~al.}{2024}]{li2024mvbench}
\begin{bchapter}
\bauthor{\bsnm{Li}, \binits{K.}},
\bauthor{\bsnm{Wang}, \binits{Y.}},
\bauthor{\bsnm{He}, \binits{Y.}},
\bauthor{\bsnm{Li}, \binits{Y.}},
\bauthor{\bsnm{Wang}, \binits{Y.}},
\bauthor{\bsnm{Liu}, \binits{Y.}},
\bauthor{\bsnm{Wang}, \binits{Z.}},
\bauthor{\bsnm{Xu}, \binits{J.}},
\bauthor{\bsnm{Chen}, \binits{G.}},
\bauthor{\bsnm{Luo}, \binits{P.}}, \betal:
\bctitle{Mvbench: A comprehensive multi-modal video understanding benchmark}.
In: \bbtitle{CVPR}
(\byear{2024})
\end{bchapter}
\endbibitem

\bibitem[\protect\citeauthoryear{Fu et~al.}{2024}]{fu2024videomme}
\begin{botherref}
\oauthor{\bsnm{Fu}, \binits{C.}},
\oauthor{\bsnm{Dai}, \binits{Y.}},
\oauthor{\bsnm{Luo}, \binits{Y.}},
\oauthor{\bsnm{Li}, \binits{L.}},
\oauthor{\bsnm{Ren}, \binits{S.}},
\oauthor{\bsnm{Zhang}, \binits{R.}},
\oauthor{\bsnm{Wang}, \binits{Z.}},
\oauthor{\bsnm{Zhou}, \binits{C.}},
\oauthor{\bsnm{Shen}, \binits{Y.}},
\oauthor{\bsnm{Zhang}, \binits{M.}}, et al.:
Video-mme: The first-ever comprehensive evaluation benchmark of multi-modal llms in video analysis.
arXiv preprint
(2024)
\end{botherref}
\endbibitem

\bibitem[\protect\citeauthoryear{Wang et~al.}{2024}]{wang2024lvbench}
\begin{botherref}
\oauthor{\bsnm{Wang}, \binits{W.}},
\oauthor{\bsnm{He}, \binits{Z.}},
\oauthor{\bsnm{Hong}, \binits{W.}},
\oauthor{\bsnm{Cheng}, \binits{Y.}},
\oauthor{\bsnm{Zhang}, \binits{X.}},
\oauthor{\bsnm{Qi}, \binits{J.}},
\oauthor{\bsnm{Gu}, \binits{X.}},
\oauthor{\bsnm{Huang}, \binits{S.}},
\oauthor{\bsnm{Xu}, \binits{B.}},
\oauthor{\bsnm{Dong}, \binits{Y.}}, et al.:
Lvbench: An extreme long video understanding benchmark.
arXiv preprint
(2024)
\end{botherref}
\endbibitem

\bibitem[\protect\citeauthoryear{Ye et~al.}{2023}]{ye2023mplug}
\begin{botherref}
\oauthor{\bsnm{Ye}, \binits{Q.}},
\oauthor{\bsnm{Xu}, \binits{H.}},
\oauthor{\bsnm{Xu}, \binits{G.}},
\oauthor{\bsnm{Ye}, \binits{J.}},
\oauthor{\bsnm{Yan}, \binits{M.}},
\oauthor{\bsnm{Zhou}, \binits{Y.}},
\oauthor{\bsnm{Wang}, \binits{J.}},
\oauthor{\bsnm{Hu}, \binits{A.}},
\oauthor{\bsnm{Shi}, \binits{P.}},
\oauthor{\bsnm{Shi}, \binits{Y.}}, et al.:
mplug-owl: Modularization empowers large language models with multimodality.
arXiv preprint
(2023)
\end{botherref}
\endbibitem

\bibitem[\protect\citeauthoryear{Chiang et~al.}{2023}]{vicuna2023}
\begin{botherref}
\oauthor{\bsnm{Chiang}, \binits{W.-L.}},
\oauthor{\bsnm{Li}, \binits{Z.}},
\oauthor{\bsnm{Lin}, \binits{Z.}},
\oauthor{\bsnm{Sheng}, \binits{Y.}},
\oauthor{\bsnm{Wu}, \binits{Z.}},
\oauthor{\bsnm{Zhang}, \binits{H.}},
\oauthor{\bsnm{Zheng}, \binits{L.}},
\oauthor{\bsnm{Zhuang}, \binits{S.}},
\oauthor{\bsnm{Zhuang}, \binits{Y.}},
\oauthor{\bsnm{Gonzalez}, \binits{J.E.}},
\oauthor{\bsnm{Stoica}, \binits{I.}},
\oauthor{\bsnm{Xing}, \binits{E.P.}}:
Vicuna: An Open-Source Chatbot Impressing GPT-4 with 90\%* ChatGPT Quality
(2023).
\url{https://lmsys.org/blog/2023-03-30-vicuna/}
\end{botherref}
\endbibitem

\end{thebibliography}

\end{document}